\documentclass{article}

 \usepackage[preprint]{neurips_2026}

\usepackage[utf8]{inputenc} 
\usepackage[T1]{fontenc}    
\usepackage{url}            
\usepackage{booktabs}       
\usepackage{amsfonts}       
\usepackage{nicefrac}       
\usepackage{microtype}      
\usepackage{xcolor}         
\usepackage{booktabs}
\usepackage{tabularx}
\usepackage{amsmath}
\usepackage{subcaption}
\usepackage{float}
\usepackage{multirow}
\usepackage{xurl}

\usepackage{textcomp}
\usepackage{listings}
\usepackage[scale=0.875]{sourcecodepro}

\usepackage{tikz}
\usetikzlibrary{shapes.multipart, positioning, shadows, fit, backgrounds, arrows.meta}

\definecolor{mediumgray}{rgb}{0.3, 0.4, 0.4}
\definecolor{mediumblue}{rgb}{0.0, 0.0, 0.8}
\definecolor{forestgreen}{rgb}{0.13, 0.55, 0.13}
\definecolor{darkviolet}{rgb}{0.58, 0.0, 0.83}
\definecolor{royalblue}{rgb}{0.25, 0.41, 0.88}
\definecolor{crimson}{rgb}{0.86, 0.8, 0.24}
\definecolor{lightyellow}{rgb}{1, 1, 0.9}

\lstdefinelanguage{JavaScript}{
  morekeywords=[1]{break, continue, delete, else, for, function, if, in,
    new, return, this, typeof, var, void, while, with},
  morekeywords=[2]{false, null, true, boolean, number, undefined,
    Array, Boolean, Date, Math, Number, String, Object},
  morekeywords=[3]{eval, parseInt, parseFloat, escape, unescape},
  sensitive,
  morecomment=[s]{/*}{*/},
  morecomment=[l]//,
  morecomment=[s]{/**}{*/}, 
  morestring=[b]',
  morestring=[b]"
}[keywords, comments, strings]

\lstdefinelanguage{JSONLD}{
  morekeywords=[1]{@context, @type, @id, @graph},
  morekeywords=[2]{croissant:Task, croissant:TaskProblem, croissant:TaskSolution, croissant:EvaluationTask, croissant:EvaluationResult, croissant:ExecutionConfig, croissant:OutputSpec, croissant:EvaluationSpec},
  sensitive=true,
  morestring=[b]",
  morestring=[b]'
}

\lstdefinelanguage{Markdown}{
  morekeywords={\#,\#\#,\#\#\#,\#\#\#\#},
  sensitive=true,
  morecomment=[s]{<!--}{-->},
}

\lstdefinestyle{JSES6Base}{
  backgroundcolor=\color{white},
  basicstyle=\ttfamily,
  breakatwhitespace=false,
  breaklines=false,
  captionpos=b,
  columns=fullflexible,
  commentstyle=\color{mediumgray}\upshape,
  emph={},
  emphstyle=\color{crimson},
  extendedchars=true,  
  fontadjust=true,
  frame=single,
  identifierstyle=\color{black},
  keepspaces=true,
  keywordstyle=\color{mediumblue},
  keywordstyle={[2]\color{darkviolet}},
  keywordstyle={[3]\color{royalblue}},
  numbers=left,
  numbersep=5pt,
  numberstyle=\tiny\color{black},
  rulecolor=\color{black},
  showlines=true,
  showspaces=false,
  showstringspaces=false,
  showtabs=false,
  stringstyle=\color{forestgreen},
  tabsize=2,
  title=\lstname,
  upquote=true  
}

\lstdefinestyle{JSONLD}{
  language=JSONLD,
  style=JSES6Base,
  frame=single,
  rulecolor=\color{lightyellow},
  framexleftmargin=5mm, 
  framexrightmargin=0mm, 
  framextopmargin=3pt,  
  framexbottommargin=3pt,  
  belowskip=-10pt,
  backgroundcolor=\color{lightyellow},
  xleftmargin=2em,
  numberstyle=\footnotesize\color{gray}\ttfamily,
  keywordstyle=\color{royalblue}, 
  keywordstyle={[2]\color{darkviolet}}, 
  stringstyle=\color{forestgreen},
  escapeinside={|}{|},
  literate={@context}{{\color{royalblue}@context}}8
           {@type}{{\color{royalblue}@type}}5
           {@id}{{\color{royalblue}@id}}3
}

\lstdefinestyle{Markdown}{
  language=Markdown,
  style=JSES6Base,
  frame=single,
  rulecolor=\color{lightyellow},
  framexleftmargin=5mm, 
  framexrightmargin=0mm, 
  framextopmargin=3pt,  
  framexbottommargin=3pt,  
  belowskip=-10pt,
  backgroundcolor=\color{lightyellow},
  xleftmargin=2em,
  numberstyle=\footnotesize\color{gray}\ttfamily,
  keywordstyle=\color{royalblue},
  stringstyle=\color{forestgreen},
  escapeinside={|}{|},
  moredelim=[is][\color{crimson}\bfseries]{**}{**},
  moredelim=[is][\color{forestgreen}\ttfamily]{`}{`},
}

\usepackage{hyperref}   
\hypersetup{
  colorlinks   = true, 
  urlcolor     = blue, 
  linkcolor    = purple, 
  citecolor    = blue 
}

\usepackage{todonotes}

\newcommand{\luis}[1]{\todo[inline,color=red!40]{#1 -- luis}}
\newcommand{\omar}[1]{\todo[inline,color=purple!40]{#1 -- Omar}}
\newcommand{\isabelle}[1]{\todo[inline,color=green!40]{#1 --Isabelle}}

\newcommand{\jon}[1]{\todo[inline,color=teal!40]{#1 -- Jon}}

\newcommand{\joaquin}[1]{\todo[inline,color=green!20]{#1 -- Joaquin}}

\newcommand{\old}[1]{}  

\title{Croissant Tasks: A Metadata Format for Reproducible Machine Learning Evaluations}

\author{%
  Omar Benjelloun\textsuperscript{1}\thanks{Authors listed in alphabetical order. Corresponding author: \href{mailto:benjello@google.com}{\texttt{benjello@google.com}}} \quad
  Leonardo Martins Bianco\textsuperscript{1} \quad
  Isabelle Guyon\textsuperscript{1,2} \\
  \textbf{Thanh Gia Hieu Khuong\textsuperscript{3}} \quad
  \textbf{Jonathan Lebensold\textsuperscript{4,5}} \\
  \textbf{Sebastian Lobentanzer\textsuperscript{6,7,8,9}} \quad
  \textbf{Luis Oala\textsuperscript{10}} \\
  \textbf{Benedictus Kent Rachmat\textsuperscript{3}} \quad
  \textbf{Ihsan Ullah\textsuperscript{2}} \quad
  \textbf{Peyman Vahidi\textsuperscript{6,7,8}} \\
  \textbf{Joaquin Vanschoren\textsuperscript{11}} \\
  \\
  \textsuperscript{1}Google DeepMind, Paris, France \quad 
  \textsuperscript{2}ChaLearn, CA, USA \\
  \textsuperscript{3}Université Paris-Saclay, Gif-sur-Yvette, France \\
  \textsuperscript{4}Jetty, Montreal, QC, Canada \\
  \textsuperscript{5}Mila, Quebec AI Institute, Montreal, QC, Canada \\
  \textsuperscript{6}Inst. of Computational Biology, Helmholtz Munich, Germany \\
  \textsuperscript{7}German Center for Diabetes Research, Munich, Germany \\ 
  \textsuperscript{8}School of CIT, TUM, Munich, Germany \quad
  \textsuperscript{9}Helmholtz AI, Munich, Germany \\
  \textsuperscript{10}Brickroad, New York, NY, USA \\
  \textsuperscript{11}Eindhoven Univ. of Technology, Eindhoven, The Netherlands
}

\begin{document}

\maketitle
\old{\isabelle{Scope -- Are we limiting ourselves to numerical experiments? Should we specify the scope of Croissant tasks (maybe it addresses and scientific or technical task, but we are presently validating on ML tasks?}}

\begin{abstract}

Reproducibility is fundamental to the scientific method, yet remains a critical challenge in machine learning. Contributing factors include underspecified execution details and brittle software environments. Human-centric remedies, such as checklists and manual verification, help but require intensive effort and fail to scale. 
To address this, we introduce \textbf{Croissant Tasks}: a declarative, \textit{machine-actionable} metadata format that abstracts low-level implementation details into high-level specifications. This format enables \textit{conceptual reproducibility}: verifying claims via independent, agent-generated implementations rather than brittle source code replication. We contribute: (1) the Croissant Tasks specification, formally decoupling task problem from solution; (2) an automated LLM pipeline that retrofits existing benchmarks into this format; and (3) empirical validation showing autonomous agents can ingest these specifications to generate functional, accurate reproduction pipelines from scratch.
We envision this format as a new foundation for automated and conceptual reproducibility in machine learning.
\end{abstract}

\vspace{0.6cm}
\section{Introduction}

Reproducibility is fundamental to the scientific method~\citep{popper1935logik}, yet it remains a critical challenge in machine learning (ML)~\citep{Baker2016-cg, wired-repro-crisis, hutson2018artificial, mit_review_reproducibility}. 
This issue is acute in benchmarks and competitions, which serve as the primary yardstick for progress and guide community research directions. Without a standardized, verifiable approach for describing and running evaluations, model comparisons across studies remain fragmented and unreliable.

This fragmentation stems from several interrelated causes: Source code is often withheld, critical execution details such as hyperparameters, data preprocessing, or evaluation prompts are underspecified, and implementations are frequently tied to specific, rapidly obsolete software environments~\citep{gundersen2022sources}. Researchers rely on ad-hoc implementations where minor differences in library versions or hardware configurations can lead to significant discrepancies in outcomes. Consequently, a substantial gap exists between a paper's high-level conceptual contribution and the brittle, low-level details required to execute it, making reproduction a labor-intensive process.

To address these issues, the community has introduced remedies targeting specific aspects of the problem. Documentation frameworks like checklists~\citep{pineau2021improving} and model cards~\citep{Mitchell_2019} enhance transparency, while reproducibility badges~\citep{ieee_reproducibility_initiative} incentivize artifact sharing. Furthermore, platforms like Hugging Face, OpenML~\citep{openml}, and Codabench~\citep{Codabench-2022} facilitate asset sharing. However, these solutions lack a formal, standardized representation of the execution flow, leaving reproduction dependent on ad-hoc setups rather than automated execution.

Recent advances in autonomous coding agents offer a paradigm shift. The difficulty of reproducing original software environments often forces the community to rely on a few fixed, legacy evaluation harnesses, leading to static and easily saturated leaderboards~\citep{liao2021are}. To address this, we propose achieving reproducibility via high-level, machine-readable specifications rather than strict access to original source code. 
We show that modern agents can interpret these specifications to synthesize functional implementations from scratch. This shifts the goalpost from \textit{technical replication} (running the exact original code) to \textit{conceptual reproducibility} (verifying the scientific claim through an independently generated implementation).
This approach not only circumvents environment dependency constraints but also provides a more rigorous test of the scientific claim by validating it through distinct, independently generated implementations.
\old{\luis{thread in d.'s ml tech debt -> ``reproducibility debt'' \cite{NIPS2015_86df7dcf}}}

To realize this vision, we introduce \textbf{Croissant Tasks}, a declarative metadata format that represents benchmarks and competitions as structured, machine-actionable data. This work builds upon the Croissant dataset standard~\citep{croissant_datasets}, which makes machine learning datasets more reusable and discoverable. 

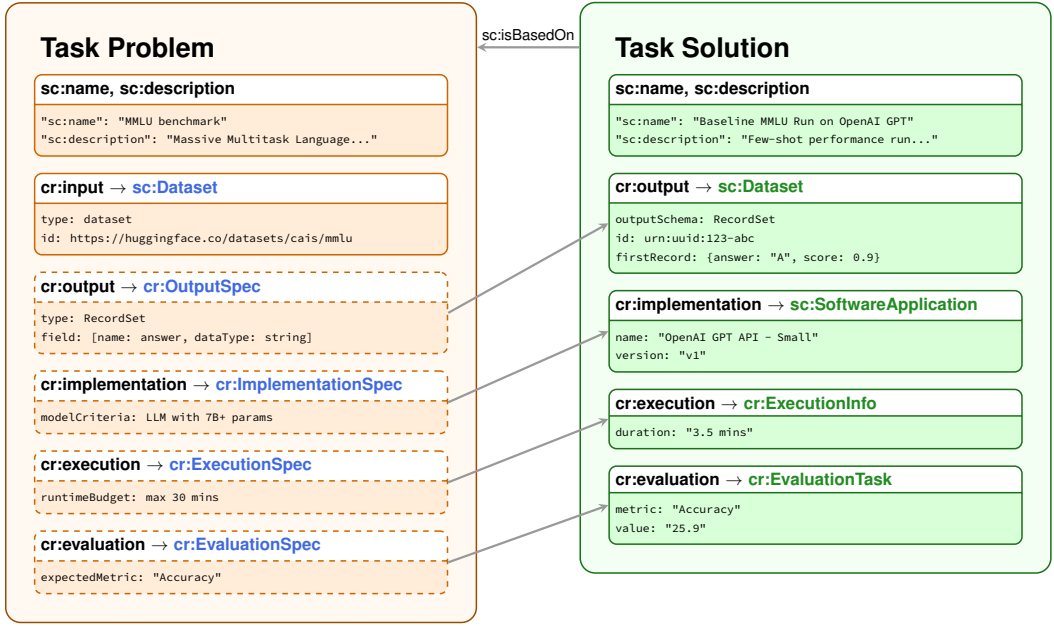
\begin{figure}[t]
    \centering
    \resizebox{\linewidth}{!}{%
        \begin{tikzpicture}[
        node distance=0.3cm and 3cm,
        basenode/.style={rectangle split, rectangle split parts=2, draw, rounded corners=4pt, text width=7.5cm, align=left, font=\sffamily\small},
        spec/.style={basenode, dashed, thick, draw=orange!80!black, rectangle split part fill={white, orange!15}},
        conc/.style={basenode, thick, solid, draw=forestgreen!80!black, rectangle split part fill={white, green!15}},
        topbox/.style={basenode, solid, thick, rectangle split part fill={white, #1}},
        arrow/.style={->, >=stealth, line width=1.2pt, draw=black!40},
        dashedarrow/.style={->, >=stealth, line width=1pt, dashed, draw=black!30}
    ]

    \node[spec, solid] (p_name) {
        \textbf{sc:name, sc:description}
        \nodepart{two}
        \begin{tabular}{@{}l@{}}
        \texttt{\scriptsize "sc:name": "MMLU benchmark"} \\
        \texttt{\scriptsize "sc:description": "Massive Multitask Language..."}
        \end{tabular}
    };
    
    \node[spec, solid, below=of p_name] (p_in) {
        \textbf{cr:input} $\rightarrow$ \textcolor{royalblue}{\textbf{sc:Dataset}}
        \nodepart{two}
        \begin{tabular}{@{}l@{}}
        \texttt{\scriptsize type: dataset} \\
        \texttt{\scriptsize id: https://huggingface.co/datasets/cais/mmlu}
        \end{tabular}
    };
    
    \node[spec, below=of p_in] (p_out) {
        \textbf{cr:output} $\rightarrow$ \textcolor{royalblue}{\textbf{cr:OutputSpec}}
        \nodepart{two}
        \begin{tabular}{@{}l@{}}
        \texttt{\scriptsize type: RecordSet} \\
        \texttt{\scriptsize field: [name: answer, dataType: string]}
        \end{tabular}
    };
    
    \node[spec, below=of p_out] (p_impl) {
        \textbf{cr:implementation} $\rightarrow$ \textcolor{royalblue}{\textbf{cr:ImplementationSpec}}
        \nodepart{two}
        \begin{tabular}{@{}l@{}}
        \texttt{\scriptsize modelCriteria: LLM with 7B+ params}
        \end{tabular}
    };
    
    \node[spec, below=of p_impl] (p_exec) {
        \textbf{cr:execution} $\rightarrow$ \textcolor{royalblue}{\textbf{cr:ExecutionSpec}}
        \nodepart{two}
        \begin{tabular}{@{}l@{}}
        \texttt{\scriptsize runtimeBudget: max 30 mins}
        \end{tabular}
    };
    
    \node[spec, below=of p_exec] (p_eval) {
        \textbf{cr:evaluation} $\rightarrow$ \textcolor{royalblue}{\textbf{cr:EvaluationSpec}}
        \nodepart{two}
        \begin{tabular}{@{}l@{}}
        \texttt{\scriptsize expectedMetric: "Accuracy"}
        \end{tabular}
    };

    \node[conc, right=of p_name] (s_name) {
        \textbf{sc:name, sc:description}
        \nodepart{two}
        \begin{tabular}{@{}l@{}}
        \texttt{\scriptsize "sc:name": "Baseline MMLU Run on OpenAI GPT"} \\
        \texttt{\scriptsize "sc:description": "Few-shot performance run..."}
        \end{tabular}
    };

    \node[conc, below=of s_name] (s_out) {
        \textbf{cr:output} $\rightarrow$ \textcolor{forestgreen}{\textbf{sc:Dataset}}
        \nodepart{two}
        \begin{tabular}{@{}l@{}}
        \texttt{\scriptsize outputSchema: RecordSet} \\
        \texttt{\scriptsize id: urn:uuid:123-abc} \\
        \texttt{\scriptsize firstRecord: \{answer: "A", score: 0.9\}}
        \end{tabular}
    };
    
    \node[conc, below=of s_out] (s_impl) {
        \textbf{cr:implementation} $\rightarrow$ \textcolor{forestgreen}{\textbf{sc:SoftwareApplication}}
        \nodepart{two}
        \begin{tabular}{@{}l@{}}
        \texttt{\scriptsize name: "OpenAI GPT API - Small"} \\
        \texttt{\scriptsize version: "v1"}
        \end{tabular}
    };
    
    \node[conc, below=of s_impl] (s_exec) {
        \textbf{cr:execution} $\rightarrow$ \textcolor{forestgreen}{\textbf{cr:ExecutionInfo}}
        \nodepart{two}
        \begin{tabular}{@{}l@{}}
        \texttt{\scriptsize duration: "3.5 mins"}
        \end{tabular}
    };
    
    \node[conc, below=of s_exec] (s_eval) {
        \textbf{cr:evaluation} $\rightarrow$ \textcolor{forestgreen}{\textbf{cr:EvaluationTask}}
        \nodepart{two}
        \begin{tabular}{@{}l@{}}
        \texttt{\scriptsize metric: "Accuracy"} \\
        \texttt{\scriptsize value: "25.9"}
        \end{tabular}
    };

    \node[yshift=0.5cm, anchor=west, font=\Large\bfseries\sffamily] at (p_name.north west) (prob_title) {Task Problem};
    \node[anchor=west, font=\Large\bfseries\sffamily] at (s_name.north west |- prob_title) (sol_title) {Task Solution};

    \begin{scope}[on background layer]
        \node[draw=orange!60!black, thick, solid, rounded corners=8pt, fill=orange!5, inner sep=15pt, fit=(prob_title) (p_name) (p_eval)] (prob_box) {};
        
        \node[draw=forestgreen!80!black, thick, solid, rounded corners=8pt, fill=green!5, inner sep=15pt, fit=(sol_title) (s_name) (s_eval)] (sol_box) {};
    \end{scope}

    \draw[arrow] (sol_box.west |- sol_title) -- node[above, font=\footnotesize\sffamily] {sc:isBasedOn} (prob_box.east |- prob_title);


    \draw[arrow] (p_out.east) -- (s_out.west);
    \draw[arrow] (p_impl.east) -- (s_impl.west);
    \draw[arrow] (p_exec.east) -- (s_exec.west);
    \draw[arrow] (p_eval.east) -- (s_eval.west);
    \end{tikzpicture}
    } 
    \vspace{1em}
    \caption{Example task problem and task solution for the MMLU benchmark.}
    \label{fig:mmlu-duality-structured}
\end{figure}

The main features of Croissant Tasks include: (1) structured task representation, defining inputs (datasets), outputs, metrics, and evaluation logic to abstract details into a higher-level specification; (2) comprehensive metadata dimensions, capturing compute environment (\textit{e.g.}, containers) and execution configurations as a blueprint for reproduction; and (3) problem-solution duality, separating the \textit{task problem} from the \textit{task solution} to enable standardized comparisons across diverse approaches (Figure~\ref{fig:mmlu-duality-structured}).

While the Croissant Tasks format is designed to be general across diverse machine learning tasks, this paper focuses specifically on evaluations and benchmarks.
We evaluate the format through two experiments: first, we show that LLMs can automatically generate Croissant Task descriptions from a diverse sample of benchmark papers; and second, we demonstrate that an autonomous agent can generate an implementation that matches published results from these generated Croissant Task descriptions.

The rest of the paper is organized as follows. Section~\ref{sec:related_work} discusses related work in ML reproducibility. Section~\ref{sec:specification} details the Croissant Tasks vocabulary and the problem/solution model. Section~\ref{sec:results} presents our experiments on bootstrapping tasks and reproducing evaluations. Section~\ref{sec:limitations} discusses limitations and future perspectives.

\section{Related Work}
\label{sec:related_work}


\paragraph{Technical replication vs. conceptual reproducibility.}
Validation efforts span a spectrum from strict technical replication to conceptual reproducibility. A systematic review found only 14\% of scores matched exactly, often due to underspecified conditions~\citep{belz2021systematic}. Literature distinguishes between reproducing methods (same code, same results) and reproducing results (reimplementation, similar results)~\citep{doi:10.1126/scitranslmed.aaf5027, pmlr-v97-bouthillier19a}, or internal and external validity~\citep{campbell1963experimental, mathison2004encyclopedia, liao2021are, hardt2026emerging, desai2025reproducibility}. We define \emph{technical replication} as executing original code in an identical environment, and \emph{conceptual reproducibility} as verifying claims via an independent implementation derived from a specification. Croissant Tasks targets the latter, by providing the formal specification needed to test claims independently.


\paragraph{Sources of irreproducibility.}
While many factors contribute to irreproducibility in machine learning generally~\citep{gundersen2022sources}, in evaluations it stems from underspecified details like prompt templates and sampling parameters~\citep{pmlr-v97-bouthillier19a, semmelrock2025reproducibility}, test case leakage~\citep{KAPOOR2023100804, swebenchleak}, and environment drift~\citep{albertoni2023reproducibilitymachinelearningterminology}. Consequently, scores frequently fail to match reported outcomes despite valid scientific claims~\citep{belz2021systematic}. Croissant Tasks addresses this by providing a machine-readable format that formalizes the evaluation, narrowing the gap between reported claims and verifiable computation.


\paragraph{Documentation artifacts: cards, checklists, and READMEs.}
Existing documentation practices occupy a spectrum from free-form prose to structured forms like model cards~\citep{Mitchell_2019}, datasheets, and reproducibility checklists~\citep{pineau2021improving}. While specific efforts like Eval Factsheets~\citep{bordes2025eval} and domain-specific cards enforce considerable structure, these artifacts remain oriented toward human readers. Their fields are populated in natural language, and their metric tables are not linked to the code or data that produced them, lacking a mechanism for automated verification. Furthermore, \citet{albertoni2023reproducibilitymachinelearningterminology} argue that releasing code and data alongside documentation is still insufficient for deep learning due to stochastic algorithms and library-level variability. Croissant Tasks complements this layer by contributing the missing verifiable artifact: a structured, machine-readable specification linking claimed outcomes to the inputs, subtasks, and metrics required to regenerate them.


\paragraph{Workflow languages and experiment tooling.}
Workflow languages like Snakemake~\citep{Mlder2025} and Nextflow~\citep{DiTommaso2017} address execution-level reproducibility, sometimes aided by automated extraction using LLMs~\citep{masera2025snakemakerseamlesslytransformingadhoc}.
In parallel, the ML ecosystem has accumulated a sprawling set of configuration frameworks, containerization tools, dependency managers, and experiment trackers~\citep{albertoni2023reproducibilitymachinelearningterminology}. Each of these addresses a slice of reproducibility, but there is no standard glue binding them to the scientific claims a paper actually makes.
Croissant Tasks sits above this layer: it specifies \emph{what} a benchmark evaluates and \emph{which} quantities constitute its outcome, leaving execution to existing mature tools.

\paragraph{Benchmarking platforms and leaderboards.}
Finally, Croissant Tasks relates to --- but differs in kind from --- existing benchmarking infrastructure. Evaluation harnesses such as \texttt{lm-evaluation-harness} \citep{eval-harness}, HELM \citep{liang2023holistic}, OpenCompass \citep{2023opencompass}, \texttt{lmms-eval} \citep{zhang2024lmmsevalrealitycheckevaluation}, \texttt{deepeval} \citep{Ip_deepeval_2026}, and Chatbot Arena \citep{chiang_chatbot_2024} provide runnable task suites, typically tied to particular modalities, model interfaces, or hosting environments. Leaderboards such as the Epoch AI benchmarking hub \citep{epoch2025benchmarkinghubupdate} and competition platforms such as Codabench~\citep{Codabench-2022} and Kaggle provide shared execution environments in which submissions are scored against fixed tasks. 
\old{\luis{todo: double check examples}}
These platforms excel at comparing methods side-by-side on a common problem, but they bind the task definition to platform-specific code and infrastructure. Croissant Tasks takes a task-centric rather than platform-centric view: its unit of description is the evaluation reported in a specific publication, comprising both the task problem and the intended solution path, expressed in a format that any harness, leaderboard, or competition platform could consume. The two approaches are complementary --- a Croissant Tasks file is precisely the kind of portable specification that such platforms could ingest to broaden the set of reproducible evaluations they host.
\section{The Croissant Tasks Vocabulary}
\label{sec:specification}

The \textbf{Croissant Tasks} vocabulary provides a declarative format to describe machine learning tasks. While the vocabulary is general enough to capture arbitrary tasks, this paper focuses on evaluation use cases, such as benchmarks, to demonstrate its benefits for automated reproducibility.

\old{\isabelle{Given the discussion above, maybe we want to distance ourselves from benchmarks and position ourselves as a means of formalizing the task described in a paper (not necessarily a benchmark paper), such that it is possible to reproduce the experiments and obtain the same conclusion.}}

\subsection{Core Concepts and JSON-LD Representation}

Croissant Tasks is built on Semantic Web technologies. It extends the widely used \url{https://schema.org} vocabulary and integrates with the Croissant Datasets vocabulary to describe input and output datasets, facilitating the reuse of existing data resources. Just like in Croissant Datasets, terms in this vocabulary use the \texttt{cr} prefix.
The central class in the specification is \texttt{cr:Task}, which describes a unit of work involving machine learning datasets and models. The main properties of a \texttt{Task} are described in Table~\ref{tab:core_properties}.

\begin{table}[htbp]
\centering
\begin{tabular}{@{}ll@{}}
\toprule
\textbf{Property} & \textbf{Description} \\
\midrule
\texttt{cr:input} & The data consumed by the task. \\ \addlinespace
\texttt{cr:output} & The data produced by the task. \\ \addlinespace
\texttt{cr:implementation} & The model, system, or code that performs the task. \\ \addlinespace
\texttt{cr:execution} & Computing environment, resources, and dependencies. \\ \addlinespace
\texttt{cr:evaluation} & Computed metrics and their values. \\ \addlinespace
\texttt{cr:subTask} & Subtasks that are part of the definition. \\
\bottomrule
\end{tabular}
\vspace{1em}
\caption{Core properties of the Croissant Tasks specification.}
\vspace{-1em}
\label{tab:core_properties}
\end{table}

\subsubsection{A Concrete Example: The MMLU Task}

We illustrate these concepts using the Massive Multitask Language Understanding (MMLU) benchmark~\citep{hendrycks2020measuring}. Figure~\ref{fig:mmlu_single_task} shows a JSON-LD snippet describing a specific MMLU evaluation run.


\begin{figure}[htbp]
\begin{lstlisting}[style=JSONLD, basicstyle=\ttfamily\scriptsize, escapeinside={(*}{*)}]
{ "@context": {
    "ex": "http://example.org/",
    "cr": "http://mlcommons.org/croissant/",
    "sc": "https://schema.org/"
  },
  "@type": "cr:Task",
  "@id": "ex:mmlu_small_fewshot",
  "sc:name": "MMLU Task - Small Model (Few-shot)",
  "cr:input": {
    "@type": "sc:Dataset",
    "@id": "https://huggingface.co/datasets/cais/mmlu",
    "sc:name": "MMLU Dataset on Hugging Face"
  },
  "cr:output": {
    "@type": "sc:Dataset",
    "@id": "urn:uuid:small-fewshot-overall-output"
  },
  "cr:implementation": {
    "@type": "sc:SoftwareApplication",
    "@id": "ex:mmlu_small_fewshot#implementation",
    "sc:name": "OpenAI GPT API - Small"
  },
  "cr:evaluation": {
    "@type": "cr:EvaluationTask",
    "@id": "ex:mmlu_evaluation_small_fewshot",
    "cr:evaluationResults": [
      {
        "@type": "cr:EvaluationResult",
        "cr:metric": "Accuracy",
        "cr:value": "25.9",
        "sc:description": "Overall Average Accuracy"
      }
    ]
  }
}
\end{lstlisting}
\caption{\texttt{cr:Task} for MMLU.}
\label{fig:mmlu_single_task}
\end{figure}

In this snippet, lines 1--5 define the context, binding \texttt{cr} and \texttt{sc} prefixes to their respective vocabularies. Line 6 declares this object to be a \texttt{cr:Task}. Lines 9--13 link to the source dataset on Hugging Face. Lines 14--17 specify where the output of the task is stored, and lines 18--22 identify the specific model used for the execution. Finally, the \texttt{cr:evaluation} property provides the evaluation results, reporting an accuracy of 25.9. Note that the \texttt{ex} domain and placeholder UUIDs are used merely for illustrative purposes.

To support complex evaluations, Croissant Tasks allows for a hierarchical structure via the \texttt{cr:subTask} property. Many benchmarks are composed of multiple subtasks or evaluation subsets. For example, the MMLU benchmark evaluates models across 57 subjects, which can be grouped into broader categories like humanities and STEM. Instead of creating a separate file for each subject, they can be represented as subtasks within the main task manifest. This structure is illustrated in Figures \ref{fig:mmlu_problem} and \ref{fig:mmlu_solution}. The top-level \texttt{cr:Task} contains a \texttt{cr:subTask} property that lists the subtasks. Each subtask is itself a task object, defining its own inputs and outputs while belonging to the overall benchmark definition.


\subsection{Decoupling Problems and Solutions}

While representing an evaluation in a single file is appropriate for self-contained tasks, benchmarks and competitions benefit from decoupling the \emph{problem specification} from the \emph{submitted solution}. This separation reflects a fundamental social aspect of the scientific method, where some researchers define a formal challenge and others propose independent solutions over time, often resulting in multiple waves of improvements against the same immutable problem. Formalizing this separation enables a standardized, machine-verifiable comparison of diverse approaches against a common requirement.

Concretely, Croissant Tasks introduces the following subclasses of \texttt{cr:Task}:
\begin{itemize}
    \item \textbf{\texttt{cr:TaskProblem}}: Represents a problem definition. It contains a mix of concrete values for the components it provides and placeholders (Specs) for the components it expects a solution to fulfill. Specs can specify requirements for the expected values. For instance, a \texttt{cr:OutputSpec} can specify a target schema for the output, while a \texttt{cr:EvaluationSpec} can define the metrics to compute.
    \item \textbf{\texttt{cr:TaskSolution}}: Represents a concrete response to a \texttt{cr:TaskProblem}. It references the problem via \texttt{sc:isBasedOn} and fills in the placeholders with actual data, code, or results that match the Spec requirements.
\end{itemize}

The core design principle is \textit{flexibility}: any component of a task, \textit{e.g.}, inputs, outputs, implementation, or evaluation, can be either provided as a concrete value in the problem or left as a placeholder for the solution to fulfill. 
%

This decoupling makes Croissant Tasks suitable for a number of important use cases:

\begin{table}[ht]
\centering
\small
\begin{tabularx}{\textwidth}{l >{\raggedright\arraybackslash}X >{\raggedright\arraybackslash}X >{\raggedright\arraybackslash}X}
\toprule
\textbf{Use Case} & \textbf{Problem Givens} & \textbf{Problem Specs} & \textbf{Solution} \\
\midrule
\textbf{Model Benchmarks} & Input Dataset & Output Schema, Model, Evaluation Metrics & Model, Output, Evaluation Results \\
\addlinespace
\textbf{Performance Benchmarks} & Code or Model, Input, Output & Evaluation Metrics & Execution Environment, Evaluation Results \\
\addlinespace
\textbf{Coding Competitions} & Input Data & Expected Output, Evaluation Metrics & Implementation, Evaluation Results \\
\bottomrule
\end{tabularx}
\vspace{1em}
\caption{Use cases enabled by decoupling problems and solutions.}
\vspace{-1em}
\label{tab:execution_patterns}
\end{table}

This separation enables the creator of a problem to publish an authoritative \texttt{cr:TaskProblem} file that defines the requirements, while others submit a lightweight \texttt{cr:TaskSolution} file that references the problem and provides the missing components. Crucially, this formal structure allows automated tools to verify that a solution ``matches'' the problem by fulfilling all specified requirements.

\paragraph{MMLU as a TaskProblem.}

Figure~\ref{fig:mmlu_problem} illustrates the MMLU benchmark described as a simplified \texttt{cr:TaskProblem}. It defines top-level specifications, and the \texttt{cr:subTask} property introduces an array of subtasks capturing specific partitions of the benchmark (\textit{e.g.}, the humanities subset). These subtasks reference the shared specifications to avoid redundancy. For instance, the humanities subtask specifies multiple inputs (preprocessed data and few-shot examples), demonstrating the format's support for complex evaluation setups.


\paragraph{Corresponding TaskSolution.}

A corresponding simplified solution file is shown in Figure~\ref{fig:mmlu_solution}. It links back to the problem via \texttt{sc:isBasedOn} and provides the concrete model and results. The solution mirrors the problem's hierarchy, with the \texttt{cr:subTask} property containing specific solutions for each subtask defined in the problem. 

\begin{figure}[htbp]
  \begin{subfigure}[t]{0.475\textwidth}
    \begin{lstlisting}[style=JSONLD, basicstyle=\ttfamily\tiny, xleftmargin=0.5em, xrightmargin=0.5em, escapeinside={(*}{*)}]
{ "@context": {...,
    "xsd": "http://www.w3.org/2001/XMLSchema#"
  },
  (*\textbf{"@type": "cr:TaskProblem"}*),
  "@id": "ex:mmlu_problem",
  "sc:name": "Massive Multitask Language Understanding",
  "cr:input": {
    "@type": "sc:Dataset",
    "@id": "https://huggingface.co/datasets/cais/mmlu"
  },
  (*\textbf{"cr:output"}*): {
    (*\textbf{"@type": "cr:OutputSpec"}*),
    "@id": "ex:mmlu#outputSpec",
    "cr:schema": {
      "@type": "cr:RecordSet",
      "cr:field": [
        {
          "sc:name": "answer",
          "cr:dataType": "xsd:string",
          "sc:valuePattern": "^[A-D]$"
        }
      ]
    }
  },
  (*\textbf{"cr:evaluation"}*): {
    (*\textbf{"@type": "cr:EvaluationSpec"}*),
    "cr:expectedMetric": ["Accuracy"]
  },
  (*\textbf{"cr:subTask"}*): [
    {
      "@type": "cr:TaskProblem",
      "@id": "ex:mmlu#humanities_fewshot",
      "sc:name": "MMLU - Humanities (Few-shot)",
      "cr:input": [
        {
          "@id": "ex:mmlu#humanities_preprocessed_spec"
        },
        {
          "@type": "sc:Dataset",
          "@id":"https://huggingface.co/datasets/cais/mmlu"
        }
      ],
      "cr:output": {
        "@id": "ex:mmlu#outputSpec"
      },
      "cr:evaluation": {
        "@id": "ex:mmlu#evaluationSpec"
      }
    },
    {
      "@type": "cr:TaskProblem",
      "@id": "ex:mmlu#stem_fewshot",
      ...
    }
  ]
}\end{lstlisting}
    \caption{MMLU benchmark defined as a \texttt{cr:TaskProblem} with subtasks.}
    \label{fig:mmlu_problem}
  \end{subfigure}
  \hfill
  \begin{subfigure}[t]{0.475\textwidth}
    \begin{lstlisting}[style=JSONLD, basicstyle=\ttfamily\tiny, xleftmargin=0.5em, xrightmargin=0.5em, escapeinside={(*}{*)}]
{ "@context": { ... },
  (*\textbf{"@type": "cr:TaskSolution"}*),
  "@id": "ex:mmlu_sol_small_fewshot",
  (*\textbf{"sc:isBasedOn"}*): {
    "@id": "ex:mmlu_problem"
  },
  "cr:implementation": {
    "@type": "sc:SoftwareApplication",
    "sc:name": "OpenAI GPT API - Small"
  },
  "cr:subTask": [
    {
      "@type": "cr:TaskSolution",
      "@id": "ex:mmlu_sol_small_fewshot#humanities_sol",
      "sc:isBasedOn": {
        "@id": "ex:mmlu#humanities_fewshot"
      },
      "cr:execution": {
        "@id": "ex:mmlu_sol_small_fewshot#execution"
      },
      "cr:implementation": {
        "@id": "ex:mmlu_sol_small_fewshot#implementation"
      },
      (*\textbf{"cr:output"}*): {
        (*\textbf{"@type": "sc:Dataset"}*),
        "@id": "urn:uuid:small-fewshot-humanities-output"
      },
      (*\textbf{"cr:evaluation"}*): {
        (*\textbf{"@type": "cr:EvaluationTask"}*),
        "cr:evaluatedSolution": {
          "@id": "ex:mmlu_sol_small_fewshot#humanities_sol"
        },
        "cr:evaluationResults": [
          {
            "cr:metric": "Accuracy",
            "cr:value": "24.4"
          }
        ]
      }
    },
    {
      "@type": "cr:TaskSolution",
      "@id": "ex:mmlu_sol_small_fewshot#stem_sol",
      ...
    }
  ],
  "cr:evaluation": {
    "@type": "cr:EvaluationTask",
    "cr:evaluationResults": [
      {
        "cr:metric": "Accuracy",
        "cr:value": "25.9"
      }
    ]
  }
}\end{lstlisting}
    \caption{A solution file pointing back to the MMLU \texttt{cr:TaskProblem} with subtask solutions.}
    \label{fig:mmlu_solution}
  \end{subfigure}
\end{figure}


\section{Empirical Validation}
\label{sec:results}

The main goals of our experiments are to evaluate (1) whether Croissant Tasks is expressive enough to represent the diversity of modern benchmarks and (2) whether Croissant Tasks enables the automated reproduction of these benchmarks. We emphasize that this early evaluation aims to demonstrate feasibility and is not intended as a comprehensive evaluation of reproducibility via Croissant Tasks. We focused on a sample of five benchmark papers accepted as orals or spotlights at the NeurIPS 2025 Datasets and Benchmarks track, covering a diverse range of modalities:\footnote{These papers are published under a CC-BY 4.0 license.}
\begin{enumerate}
    \item \textit{Absence Bench: Language Models Can’t See What’s Missing}~\citep{fu2026absence}
    \item \textit{CoRe: Benchmarking LLMs’ Code Reasoning Capabilities through Static Analysis Tasks}~\citep{xie2026core}
    \item \textit{MedSG-Bench: A Benchmark for Medical Image Sequences Grounding}~\citep{yue2026medsgbench}
    \item \textit{NOVA: A Benchmark for Rare Anomaly Localization and Clinical Reasoning in Brain MRI}~\citep{bercea2026nova}
    \item \textit{SAGE-Eval: Evaluating LLMs for Systematic Generalizations of Safety Facts}~\citep{yueh-han2026sageeval}
\end{enumerate}

To address these questions, we use an LLM in an agentic setup to (1) generate Croissant Tasks descriptions from these papers and (2) produce implementations from the generated Croissant Tasks descriptions. We evaluate the results via a combination of automated validation and human expert review. Appendix \ref{app:details_exp} provides details on the setup and models used.

The resulting Croissant Tasks files and generated code can be found in our \href{https://github.com/mlcommons/croissant/tree/main/tasks}{GitHub repository}\footnote{GitHub repository: \url{https://github.com/mlcommons/croissant/tree/main/tasks}}. 

\subsection{Expressivity and Automated Extraction of Croissant Tasks Files}\label{sec:expressivity}
We assess the expressivity of our proposed format by generating Croissant Tasks descriptions for each of the research papers above.
Guided by the skill file specified in Appendix~\ref{app:skills-paper2ct}, we instruct an autonomous agent to generate the corresponding Croissant Task description. We give the agent access to the language specification and tools to validate generated files.

This experiment also evaluates the extent to which Croissant Tasks description can be automatically obtained from research papers via LLM agents. The quality of the generated descriptions is evaluated in two-steps: (1) automated validation against SHACL shapes using our Python library; and (2) expert human review, verifying alignment between the generated metadata and the source publication to ensure a high-fidelity assessment of accuracy and completeness. The results are displayed in Table~\ref{tab:field_coverage}.

\begin{table}[htbp]
\centering
\begin{tabular}{@{}lcccccc@{}}
\toprule
 & \textbf{Absence} & \textbf{CoRe} & \textbf{MedSG} & \textbf{NOVA} & \textbf{SAGE} & \textbf{Average} \\ \midrule
\textbf{Coverage (\%)} & 100 & 87 & 100 & 100 & 100 & 97.4 \\ \bottomrule
\end{tabular}%
\vspace{1em}
\caption{Percentage of fields successfully covered by the generated Croissant Tasks files across the evaluated papers, after the agent's first attempt.}
\vspace{-1em}
\label{tab:field_coverage}
\end{table}


The results above show that Croissant Tasks is expressive enough to represent the evaluated benchmarks. The 87\% score on CoRe is because the extraction missed three hyperparameters described in the paper, so it's a failure of the extraction process rather than the format itself.

These results show that LLM agents can reliably extract Croissant Tasks files from research papers, lowering the adoption barrier for researchers who can generate formal specifications from their manuscripts without learning the JSON-LD vocabulary. It also allows retroactively formalizing past literature, unlocking the potential to create large-scale machine-readable benchmark databases supporting advanced search, cross-study analysis, and automated ingestion into evaluation platforms.

\subsection{Agentic Code Generation for Automated Reproduction}\label{sec:agentic}

We evaluate the feasibility of conceptual reproducibility by instructing an autonomous coding agent to generate implementations from scratch, using the Croissant Task descriptions and the skill specified in Appendix~\ref{app:skills-ct2code}. Because of resource limitations, for each paper we attempt to reproduce only one of the reported model baselines. Evaluated models included API-based models like Gemini and Claude, as well as locally-run open-weight models like Qwen. 

We compare code generation under two distinct settings to assess the sufficiency of our format: (1) providing the research paper PDF (baseline), and (2) providing the Croissant Task file. We strictly monitored web access to ensure the agent does not retrieve the paper's own code during the implementation process. 

We measure the percentage of evaluation metrics that the agent was able to implement correctly under these conditions, as determined by expert human verification. We allowed minor numerical discrepancies, due to external factors like model version changes or hardware non-determinism. Table~\ref{tab:metrics_reproduction} summarizes these results, and the details for each paper are available in Appendix~\ref{app:details_exp}.

\begin{table}[htbp]
\centering
\begin{tabular}{@{}lcccccc@{}}
\toprule
 \textbf{Setting} & \textbf{Absence} & \textbf{CoRe} & \textbf{MedSG} & \textbf{NOVA} & \textbf{SAGE} & \textbf{Average} \\ \midrule
Croissant Tasks Only (\%) & 100 & 100 & 100 & 85.7 & 100 & 97.1 \\
PDF Only (\%)       & 100 & 100 & 100 & 50 & 100 & 90 \\ \bottomrule
\end{tabular}%
\vspace{1em}
\caption{Percentage of evaluation metrics correctly implemented by the agent under two different information settings. In both cases, the interaction was limited to at most 3 human guidance prompts.}
\label{tab:metrics_reproduction}
\vspace{-1em}
\end{table}

These results show high success rates for both the baseline and Croissant Tasks-only settings. Details on metric proximity to ground truth and the required human prompts are in Appendix~\ref{app:details_exp}.

The complex NOVA benchmark resulted in a lower success rate, revealing two distinct failure modes. First, as noted by~\citet{starace2025paperbench}, the agent occasionally terminated prematurely when processing full PDFs due to context overload, omitting metrics. This issue did not occur with structured Croissant Tasks, where the compact specification provided a clear blueprint. Second, the agent sometimes substituted complex evaluation logic with simplified proxies to avoid algorithmic complexity. This affected both settings, but could be mitigated in Croissant Tasks by providing more detailed metric descriptions.

These results demonstrate that providing the agent exclusively with Croissant Tasks files is sufficient for automated reproduction, yielding successful implementations while significantly reducing context window consumption.

\section{Discussion and Future Work}
\label{sec:limitations}

\subsection{Other Benefits of Croissant Tasks}

While we focused on reproducibility in this paper, Croissant Tasks also offers several other benefits:

\paragraph{Interoperability.}
By providing a standardized, declarative representation of tasks based on Web standards, Croissant Tasks enables interoperability across several dimensions. First, it facilitates platform and tool interoperability, allowing a benchmark or competition defined on one platform (e.g., Codabench~\citep{Codabench-2022}) to be seamlessly imported and executed on another. This decouples the benchmark definition from the specific software used to run it, enabling diverse evaluation frameworks like HELM or \texttt{lm-evaluation-harness} to ingest the same task specification. Second, the formal definition of inputs and outputs enables model and benchmark interoperability by supporting the ``hot-swapping'' of components; researchers can easily use a new model in an existing task without writing custom adapter code. Finally, integration with the Croissant Datasets vocabulary ensures data interoperability with the broader data ecosystem, allowing any system capable of parsing Croissant datasets to understand the required data lineage and structures.

\paragraph{Evolution and Reuse.}
Separating problem definitions and solutions simplifies both reuse and task evolution. First, it enables structured reuse: the \texttt{cr:TaskProblem} and \texttt{cr:TaskSolution} constructs allow testing new methods on existing tasks without changing the evaluation protocol. Smaller components, like metrics or preprocessing steps, can also be extracted and reused. Second, it supports evolution and adaptation. Tasks can be versioned as they grow (e.g., adding new metrics). Furthermore, because the format is declarative rather than procedural, it adapts to software changes; if a library becomes obsolete, an agent can use the specification to generate a new implementation.

\paragraph{Discoverability.}
By extending the \url{https://schema.org} vocabulary, Croissant Tasks inherits its benefits for web discoverability. Similar to the impact of the Croissant Datasets vocabulary, this standardization enables the creation of custom search engines and centralized indexes. This allows both human researchers and autonomous agents to discover relevant tasks, benchmarks, or evaluation protocols more easily, using structured queries about inputs, outputs, or expected metrics.

\subsection{Limitations and Challenges}

A number of factors need to be tackled to realize the full potential of Croissant Tasks:

\paragraph{Adoption and Ease of Use.}
To become a community standard, Croissant Tasks must overcome several adoption challenges. First, to minimize overhead, creating specifications must not burden researchers. While LLM agents can automate generation, robust visualizers, editors, and validation tools are needed for ease of use. Second, institutional incentives are needed; adoption likely requires mandates from top-tier conferences, such as adding Croissant Task files to reproducibility checklists. Finally, achieving scale requires platform integration. Mirroring Croissant Datasets' adoption by platforms like Hugging Face and Kaggle, we envision benchmark platforms natively hosting and executing Croissant Tasks.

\paragraph{Fidelity and Quality of Metadata.}
Reproducibility and other benefits depend heavily on metadata fidelity and completeness. A task file must accurately capture all critical details, such as specific data splits, evaluation settings, and environment requirements. To ensure this completeness without adding manual burden, Croissant Tasks can be integrated with experiment tracking libraries (e.g., \href{https://wandb.ai/site/}{Weights \& Biases}) to automatically capture the required metadata as the researcher is developing their implementation. High-fidelity metadata is also essential for defining the bounds of functional reproducibility; since we target functional equivalence rather than exact replication, precise metadata is required to set acceptable tolerances for verification and to enable the reliable reuse of components.

\paragraph{Generalization and the Vision of Reproducible Science.}
While this work focuses specifically on benchmarks and evaluations, the long-term vision of Croissant Tasks is to encompass the full spectrum of machine learning tasks. Benchmarks are relatively simple to model because they typically involve fixed datasets and static evaluation protocols. Other tasks, such as training pipelines, data augmentation strategies, or complex multi-stage workflows, present significantly higher complexity. We do not aim to prescribe rigid representations for all possible tasks; rather, we believe that conventions for describing them should emerge organically within their respective research communities. By providing a foundation for machine-actionable task descriptions, we hope to contribute to the broader vision of reproducible science. Shifting the focus from brittle technical replication to verifiable, high-level specifications will allow the community to accelerate progress in a principled and trusted way.
\section{Conclusion}
\label{sec:conclusion}

In this paper, we presented Croissant Tasks, a metadata format designed to enable conceptual reproducibility in machine learning by representing evaluations as structured, machine-actionable data. We demonstrated its potential through agentic flows that (1) extract task descriptions from benchmark papers and (2) generate the necessary artifacts to implement and run these evaluations, achieving a high success rate. To realize the full potential of this approach, future work will focus on further refining the agentic workflows for more robust extraction and implementation, conducting thorough evaluations across a broader set of benchmarks, and extending it to a wider spectrum of machine learning tasks.

\bibliographystyle{plainnat}
\bibliography{references}

\appendix

\section{Croissant Tasks Turtle Schema}
\label{app:schema}

The following listing provides the formal schema definition for Croissant Tasks in Turtle (.ttl) format, as extracted from the active specification proposal. Note that while the schema uses the \texttt{croissant:} and \texttt{schema:} prefixes, the examples in the main text use \texttt{cr:} and \texttt{sc:} respectively for brevity. Both mappings point to the same base URIs.

\begin{lstlisting}[basicstyle=\ttfamily\scriptsize, frame=single, breaklines=true]
@prefix rdf: <http://www.w3.org/1999/02/22-rdf-syntax-ns#> .
@prefix rdfs: <http://www.w3.org/2000/01/rdf-schema#> .
@prefix schema: <https://schema.org/> .
@prefix croissant: <http://mlcommons.org/croissant/> .

# Top-level classes

croissant:Task a rdf:Class ;
  rdfs:label "Task" ;
  rdfs:comment "A generic task class for Croissant." ;
  rdfs:subClassOf schema:CreativeWork .

croissant:TaskProblem a rdf:Class ;
  rdfs:label "TaskProblem" ;
  rdfs:comment "A TaskProblem defines a 'problem' to be solved, i.e., an incomplete task. It uses a mix of concrete properties for components it provides ('givens') and Spec properties for components it expects a solution to provide. A TaskProblem inherits all properties from cr:Task. The role of each property is determined by the type of its value: Given Component: The property's value is a concrete type (e.g., cr:input is assigned a schema:Dataset). Expected Component: The property's value is a Spec type (e.g., cr:output is assigned a cr:OutputSpec)." ;
  rdfs:subClassOf croissant:Task .

croissant:TaskSolution a rdf:Class ;
  rdfs:label "TaskSolution" ;
  rdfs:comment "A TaskSolution represents a specific, concrete answer to a TaskProblem. It provides the actual values for the components that were marked as Specs in the problem definition. A TaskSolution inherits all properties from cr:Task to describe the concrete components of the solution." ;
  rdfs:subClassOf croissant:Task .

croissant:EvaluationTask a rdf:Class ;
  rdfs:label "EvaluationTask" ;
  rdfs:comment "A specialization of cr:Task that includes evaluation-specific information, such as metrics and results." ;
  rdfs:subClassOf croissant:Task .

croissant:EvaluationResult a rdf:Class ;
  rdfs:label "EvaluationResult" ;
  rdfs:comment "A structured result of an evaluation, pairing a metric with its value." .

# Task properties

croissant:input a rdf:Property ;
  rdfs:label "input" ;
  rdfs:comment "The primary data used as input for the task, typically provided as a Croissant dataset. This field can be repeated in case of multiple inputs." ;
  schema:domainIncludes croissant:Task ;  
  schema:rangeIncludes croissant:Dataset, schema:Dataset, schema:URL, croissant:InputSpec .

croissant:output a rdf:Property ;
  rdfs:label "output" ;
  rdfs:comment "The data generated by completing the task, typically provided as a Croissant dataset. This field can be repeated in case of multiple output datasets." ;
  schema:domainIncludes croissant:Task ;  
  schema:rangeIncludes schema:Dataset, schema:SoftwareSourceCode, croissant:OutputSpec .

croissant:implementation a rdf:Property ;
  rdfs:label "implementation" ;
  rdfs:comment "The specific program that executes the task. This can be source code, or an executable representation, such as a docker container." ;
  schema:domainIncludes croissant:Task ;  
  schema:rangeIncludes schema:SoftwareApplication, schema:SoftwareSourceCode, croissant:ImplementationSpec .

croissant:execution a rdf:Property ;
  rdfs:label "execution" ;
  rdfs:comment "Information about the execution of the task." ;
  schema:domainIncludes croissant:Task ;  
  schema:rangeIncludes croissant:ExecutionInfo, croissant:ExecutionSpec .

croissant:evaluation a rdf:Property ;
  rdfs:label "evaluation" ;
  rdfs:comment "The evaluation of the task, represented as another task." ;
  schema:domainIncludes croissant:Task ;  
  schema:rangeIncludes croissant:EvaluationTask, croissant:EvaluationSpec .

croissant:subTask a rdf:Property ;
  rdfs:label "subTask" ;
  rdfs:comment "One or more subtasks that are part of the definition of this task." ;
  schema:domainIncludes croissant:Task ;  
  schema:rangeIncludes croissant:Task .

# Evaluation properties

croissant:evaluationResults a rdf:Property ;
  rdfs:label "evaluationResults" ;
  rdfs:comment "The results of the evaluation, represented as instances of cr:EvaluationResult." ;
  schema:domainIncludes croissant:EvaluationTask ;
  schema:rangeIncludes croissant:EvaluationResult .

croissant:evaluatedSolution a rdf:Property ;
  rdfs:label "evaluatedSolution" ;
  rdfs:comment "The TaskSolution that was evaluated." ;
  schema:domainIncludes croissant:EvaluationTask ;
  schema:rangeIncludes croissant:TaskSolution .

croissant:metric a rdf:Property ;
  rdfs:label "metric" ;
  rdfs:comment "The metric used for a specific result." ;
  schema:domainIncludes croissant:EvaluationResult ;
  schema:rangeIncludes schema:Text, schema:URL .

croissant:expectedMetric a rdf:Property ;
  rdfs:label "expectedMetric" ;
  rdfs:comment "The metric expected to be calculated for this task." ;
  schema:domainIncludes croissant:EvaluationSpec ;
  schema:rangeIncludes schema:Text, schema:URL .

croissant:value a rdf:Property ;
  rdfs:label "value" ;
  rdfs:comment "The value of the result." ;
  schema:domainIncludes croissant:EvaluationResult ;
  schema:rangeIncludes schema:QuantitativeValue, schema:Text, schema:Number .

# Reusing schema:isBasedOn to link a TaskSolution or EvaluationTask to the TaskProblem it addresses.
schema:isBasedOn schema:domainIncludes croissant:TaskSolution, croissant:EvaluationTask ;
  schema:rangeIncludes schema:URL, croissant:TaskProblem ;
  rdfs:comment "A reference to the @id of the cr:TaskProblem that this solution or evaluation addresses." .

# Spec Classes

croissant:InputSpec a rdf:Class ;
  rdfs:label "InputSpec" ;
  rdfs:comment "Specifies the requirements for an input that a solution must provide. This is useful for 'bring your own data' style tasks." ;
  rdfs:subClassOf schema:Dataset .

croissant:OutputSpec a rdf:Class ;
  rdfs:label "OutputSpec" ;
  rdfs:comment "Specifies the requirements for the output that a solution must generate." ;
  rdfs:subClassOf schema:Dataset .

croissant:ImplementationSpec a rdf:Class ;
  rdfs:label "ImplementationSpec" ;
  rdfs:comment "Specifies the technical requirements for a solution's implementation (e.g., code, model)." ;
  rdfs:subClassOf schema:SoftwareApplication .

croissant:ExecutionSpec a rdf:Class ;
  rdfs:label "ExecutionSpec" ;
  rdfs:comment "A placeholder for execution information that has not yet been logged because an experiment has not yet been run." .

croissant:EvaluationSpec a rdf:Class ;
  rdfs:label "EvaluationSpec" ;
  rdfs:comment "A placeholder for evaluation metrics that have not been specified for that task." .

# Spec properties

croissant:schema a rdf:Property ;
  rdfs:label "schema" ;
  rdfs:comment "A formal Croissant schema (cr:RecordSet) defining the required structure of the dataset (input or output)." ;
  schema:domainIncludes croissant:InputSpec, croissant:OutputSpec ;
  schema:rangeIncludes croissant:RecordSet .

croissant:tests a rdf:Property ;
  rdfs:label "tests" ;
  rdfs:comment "A test or suite of tests (cr:Test or schema:URL) that the implementation must pass." ;
  schema:domainIncludes croissant:ImplementationSpec ;
  schema:rangeIncludes croissant:Test, schema:URL .

croissant:environment a rdf:Property ;
  rdfs:label "environment" ;
  rdfs:comment "A description of or link to the required execution environment (e.g., a requirements.txt file)." ;
  schema:domainIncludes croissant:ImplementationSpec ;
  schema:rangeIncludes schema:CreativeWork, schema:URL .

# Test Classes

croissant:Test a rdf:Class ;
  rdfs:label "Test" ;
  rdfs:comment "Represents a test or a test suite for verifying an implementation." .

# Test properties

croissant:testCommand a rdf:Property ;
  rdfs:label "testCommand" ;
  rdfs:comment "The command to execute the test." ;
  schema:domainIncludes croissant:Test ;
  schema:rangeIncludes schema:Text .

# Execution Classes

croissant:ExecutionInfo a rdf:Class ;
  rdfs:label "ExecutionInfo" ;
  rdfs:comment "Information about the execution of a task." .

croissant:ExecutionConfig a rdf:Class ;
  rdfs:label "ExecutionConfig" ;
  rdfs:comment "Configuration used for execution, such as hyperparameters." ;
  rdfs:subClassOf croissant:ExecutionInfo .

croissant:ExecutionTrace a rdf:Class ;
  rdfs:label "ExecutionTrace" ;
  rdfs:comment "Trace information from execution, such as logs and metrics." ;
  rdfs:subClassOf croissant:ExecutionInfo .

# Execution properties

croissant:hyperparameter a rdf:Property ;
  rdfs:label "hyperparameter" ;
  rdfs:comment "A hyperparameter used in the execution configuration." ;
  schema:domainIncludes croissant:ExecutionConfig ;
  schema:rangeIncludes schema:PropertyValue .

# External properties reused or formalized here

schema:valuePattern a rdf:Property ;
  rdfs:label "valuePattern" ;
  rdfs:comment "A regular expression pattern to validate field values." ;
  schema:domainIncludes croissant:Field ;
  schema:rangeIncludes schema:Text .
\end{lstlisting}

\old{
\section{Evaluation rubric for generated Croissant Tasks files}
\label{app:paper2ct-rubric}

The following rubric was used to evaluate the quality of the Croissant Tasks files generated from each paper's PDF.

\begin{table}[htbp]
\centering
\label{tab:evaluation_rubric}
\resizebox{\textwidth}{!}{%
\begin{tabular}{@{}lp{4cm}p{4.5cm}p{4.5cm}@{}}
\toprule
\textbf{Criterion} & \textbf{Description} & \textbf{Score 5 (Optimal)} & \textbf{Score 1 (Poor)} \\ \midrule
\textit{Completeness} & Inclusion of all fundamental benchmark elements. & All \texttt{subTasks}, \texttt{input} datasets, \texttt{output} schemas, and metrics are mapped without omissions. & Crucial elements are missing, rendering the benchmark uninterpretable. \\ \addlinespace
\textit{Semantic Correctness} & Adherence to the formal Croissant and \texttt{schema.org} ontologies. & Perfect application of classes and primitive types respecting concept boundaries, passes validation. & Incorrect class usage that structurally violates the underlying RDF graph logic. \\ \addlinespace
\textit{Actionability} & Sufficiency of information for automated code generation. & Fully specified resources and unambiguous metric declarations. & The metadata is too vague to guide an autonomous agent during implementation. \\ \bottomrule
\end{tabular}%
}
\vspace{1em}
\caption{Human Evaluation Rubric for Croissant Task Specifications.}
\end{table}
}

\section{Skill File and Prompt for Agentic Writing of Croissant Tasks}
\label{app:skills-paper2ct}
We propose the following \texttt{SKILL.md} file to assist the agent to transform a benchmark paper into a Croissant Tasks specification.
\begin{lstlisting}[basicstyle=\ttfamily\scriptsize, frame=single, breaklines=true]
---
name: pdf2ct
description: >-
  Converts academic papers (PDFs) describing machine learning benchmarks or tasks into MLCommons Croissant Tasks JSON-LD files. Use when you need to extract task definitions, inputs, outputs, evaluation metrics, and baseline results from a paper and represent them in Croissant Tasks format. Don't use for general PDF text extraction or dataset metadata (use Croissant Datasets for that).
---

# PDF to MLCommons Croissant **Tasks** - Agent Runbook

## Objective

You are given an academic paper (PDF) that introduces a machine learning **benchmark or task** (e.g. MMLU, AbsenceBench, GSM8K). Your job is to read the paper, extract the task definition and any evaluated baselines, and produce valid **MLCommons Croissant Tasks** JSON-LD files:

1. Exactly one **`TaskProblem`** describing the benchmark abstractly (inputs, expected outputs, evaluation metrics, subtasks).
2. Zero or more **`TaskSolution`** files - one per concrete model/approach the paper evaluates (with hyperparameters and `EvaluationResult`s).

Validate every file with the Croissant Tasks SHACL validator (`pyshacl` + the shapes/ontology from the latest commit in PR #1017), iterate up to 3 rounds to fix validation errors, and write an executive summary + validation report.

Croissant Tasks is distinct from Croissant Datasets. The agent emits **task descriptions**, not dataset metadata. If the paper is purely a dataset (no defined task/benchmark), flag that in the summary and still emit the best possible `TaskProblem` using `InputSpec`.

---

## Croissant Mapping Guidance

- Use `croissant:TaskProblem` to define a benchmark or a component of it (subtask) abstractly. It describes the expected inputs, outputs, and evaluation criteria, serving as a blueprint for solutions.
- Use `croissant:Task` as a base class or when a single file represents both the problem definition and a particular solution without making a distinction between the two. It can also represent a benchmark suite or collection of related targets.
- Use `croissant:subTask` to break down a task into smaller, meaningful components. In a `TaskProblem`, it describes a sub-problem (e.g., a specific dataset split or domain). In a `TaskSolution`, it prescribes the specific solution to that sub-problem.
- Do not model one implementation's internal pipeline as normative task structure unless those intermediate artifacts are themselves benchmarked as standalone targets.
- Put datasets, schemas, fixed examples, and reference context in `croissant:input` when they are part of the task definition.
- In a `TaskProblem`, use `croissant:output` with an `OutputSpec` to describe the expected shape and data type of the task's results. In a `TaskSolution`, use `croissant:output` to point to the concrete dataset (or its UUID) containing the actual results produced by the implementation.
- Use `croissant:EvaluationSpec` to name expected metrics, but keep detailed scoring rules in companion documentation when they are more specific than the ontology can express.
- Use `croissant:implementation` to point to reference code or systems when that helps orientation, but treat those implementations as descriptive rather than normative unless the benchmark explicitly constrains them.
- Use `croissant:TaskSolution` to document a concrete run of a model or approach on the task. Each row of a results table in a paper typically corresponds to a `TaskSolution` (a specific model evaluated on various subtasks). It is optional if the user only wants to define the problem.

---

## REQUIRED OUTPUT FILES (MANDATORY)

**You MUST write all of the following files to `{{results_dir}}`.
The task is NOT complete until every file exists and is non-empty. No exceptions.**

| File | Description |
|------|-------------|
| `{{results_dir}}/problem.jsonld` | The `croissant:TaskProblem` JSON-LD for the benchmark |
| `{{results_dir}}/solutions/<slug>.jsonld` | One `croissant:TaskSolution` per evaluated baseline (e.g. `gpt4_fewshot.jsonld`). May be zero files if the paper reports no baselines. |
| `{{results_dir}}/summary.md` | Executive summary - what was extracted, inferred, skipped |
| `{{results_dir}}/validation_report.json` | Structured validation results with `stages`, `results`, `overall_passed` |

If you finish your analysis but have not written all files, go back and write them before stopping.

---

## Parameters

| Parameter | Template Variable | Default | Description |
|-----------|------------------|---------|-------------|
| Results directory | `{{results_dir}}` | `/app/results` (Jetty) / `./results` (local) | Output directory for all results |
| PDF location | `{{pdf_location}}` | - | The location of the PDF to convert (local path or URL), specified in the user prompt. |
| Paper URL | `{{paper_url}}` | (empty) | Optional arXiv/DOI URL - used to derive stable `@id` base IRIs |
| Dataset URL | `{{dataset_url}}` | (empty) | Optional canonical dataset URL (HuggingFace, GitHub) used as the task's `croissant:input` |
| Croissant Tasks README | - | `README.md` at the latest commit in PR #1017 | Primary readable description of the specification that you should use as principle. |

---

## Dependencies

| Dependency | Type | Required | Description |
|------------|------|----------|-------------|
| pyshacl | Python package | Yes | SHACL validator engine |
| rdflib | Python package | Yes | RDF graph parser (pulls in JSON-LD support) |
| Croissant Tasks shapes TTL | File | Yes | `croissant-tasks-shapes.ttl` from the latest commit in PR #1017 |
| Croissant Tasks ontology TTL | File | Yes | `croissant-tasks.ttl` from the latest commit in PR #1017 |
| Python Validator | File | Yes | Python validator script available in the latest commit of PR #1017 (e.g., `validator.py`) |

Use the latest commit in PR #1017.

---

## Step 1: Environment Setup

```bash
# Install SHACL validator and RDF toolkit
pip install pyshacl rdflib

# Output directories
mkdir -p {{results_dir}}/solutions

# Download the Croissant Tasks shapes + ontology at the pinned commit
# Use the latest commit in PR #1017
COMMIT=<latest_commit_in_PR_1017>
BASE=https://raw.githubusercontent.com/mlcommons/croissant/${COMMIT}/tasks
curl -fsSL "${BASE}/croissant-tasks-shapes.ttl" -o {{results_dir}}/croissant-tasks-shapes.ttl
curl -fsSL "${BASE}/croissant-tasks.ttl"        -o {{results_dir}}/croissant-tasks.ttl

# Verify the PDF exists (local path) or download it if it's a URL
# The location will be specified in the user prompt.
# Example for local file: ls -la {{pdf_location}}
# Example for URL: curl -fsSL {{pdf_location}} -o paper.pdf
```

Verify all required inputs and dependency files are present before proceeding.

---

## Step 2: Read and Analyze the Paper

Read the PDF in full. A benchmark paper usually defines a task + evaluates some baselines. Extract:

### Task Identity
- **Name** - official benchmark name (e.g. "MMLU", "AbsenceBench")
- **Description** - 1-3 sentence summary of what a solution to the task must do
- **Paper URL** - arXiv/DOI link (fall back to `{{paper_url}}` if unknown)
- **Task homepage / repo** - GitHub, leaderboard, or project site if mentioned

### Inputs
- **Dataset used as input** - URL (HuggingFace, GitHub, paper supplementary). If the paper expects BYO data ("bring your own"), use `croissant:InputSpec` instead of a concrete `schema:Dataset`.
- **Input schema** - what fields a datum has (question text, context, images, etc.)

### Outputs
- **Expected output fields** - exact schema: name, data type (xsd:string / xsd:float / xsd:integer / xsd:boolean), description, and any value constraints (e.g. regex `^[A-D]$` for multiple-choice).
- **Scalar vs. structured** - is each prediction a single value, a vector, a classification label, or structured?

### Evaluation
- **Metrics** - the primary and secondary metrics (e.g. "Accuracy", "F1-Score", "Calibration Error", "Exact Match").
- **Evaluation protocol** - few-shot / zero-shot / chain-of-thought / fine-tuned.

### Subtasks
- Does the benchmark split into sub-categories (MMLU humanities/STEM/social/other, GLUE tasks)? Each is a nested `croissant:subTask` under `croissant:TaskProblem`.

### Baselines Reported
For every model the paper evaluates, capture:
- **Model name & provider** (e.g. "OpenAI GPT-4", "Google Gemini 3")
- **Hyperparameters** - temperature, top-p, max tokens, few-shot k, etc. (see paper's appendix / experimental setup)
- **Results per metric per subtask** - the numbers in the main results table
- **Overall score** - if reported

**Important**: Distinguish explicitly stated information from inferred information. Track the distinction - you will report it in the summary.

---

## Step 3: Derive the `@id` base

Pick a stable, de-reffable `@id` base for the JSON-LD documents. Order of preference:

1. `{{paper_url}}` (e.g. `https://arxiv.org/abs/2212.xxxxx#<slug>`)
2. Canonical task homepage
3. Fallback: `http://example.org/<kebab-case-slug>` (following the MMLU example's convention)

Use a single base throughout - every `@id` should be consistent and unique. For the problem and its subtasks, use fragments: `<base>#<subtask_slug>`. For solutions, use `<base>_solution_<model_slug>`.

---

## Step 4: Build the `TaskProblem` JSON-LD

Write the problem to `{{results_dir}}/problem.jsonld`.

### `@context` (copy verbatim)

```json
"@context": {
  "croissant": "http://mlcommons.org/croissant/",
  "schema":    "https://schema.org/",
  "xsd":       "http://www.w3.org/2001/XMLSchema#"
}
```

### Required structure

```json
{
  "@context": { "...": "see above" },
  "@type": "croissant:TaskProblem",
  "@id":   "<base>",
  "schema:name":        "Benchmark Name",
  "schema:description": "What a solution to this task must do.",

  "croissant:input":  { /* schema:Dataset OR croissant:InputSpec */ },
  "croissant:output": { /* croissant:OutputSpec (REQUIRED to make it a Problem) */ },
  "croissant:implementation": { /* optional: croissant:ImplementationSpec */ },
  "croissant:execution": { /* optional: croissant:ExecutionSpec */ },

  "croissant:evaluation": { /* croissant:EvaluationSpec with expectedMetric */ },

  "croissant:subTask": [ /* optional: nested TaskProblems for sub-categories */ ]
}
```

### SHACL constraints you MUST satisfy (from `croissant-tasks-shapes.ttl`)

**TaskShape (base)** - applies to every Task/TaskProblem/TaskSolution/EvaluationTask:
- `croissant:input` --> `croissant:Dataset` | `schema:Dataset` | URL (IRI) | `croissant:InputSpec`
- `croissant:output` --> `schema:Dataset` | `schema:SoftwareSourceCode` | `croissant:OutputSpec`
- `croissant:implementation` --> `schema:SoftwareApplication` | `schema:SoftwareSourceCode` | `croissant:ImplementationSpec`
- `croissant:execution` --> `croissant:ExecutionInfo` | `croissant:ExecutionConfig` | `croissant:ExecutionTrace` (NB: for TaskProblem, it must specifically be `croissant:ExecutionSpec`)
- `croissant:evaluation` on a **Problem** --> `croissant:EvaluationTask` | `croissant:EvaluationSpec`
- `croissant:subTask` --> must be a `croissant:Task` (or subclass: `TaskProblem` / `TaskSolution` / `EvaluationTask`)

**TaskProblemShape** - adds:
- MUST have at least ONE of: `input` as `InputSpec`, `output` as `OutputSpec`, or `implementation` as `ImplementationSpec`.
  *Typical benchmark papers satisfy this via `OutputSpec`.*
- If `execution` is present --> must be `croissant:ExecutionSpec`.
- If `evaluation` is present --> must be `EvaluationTask` or `EvaluationSpec`.

**OutputSpec** must contain `croissant:schema` --> a `croissant:RecordSet` with >= 1 `croissant:field`.

### Output schema patterns

**(a) Scalar** - single float/number per prediction:
```json
"croissant:output": {
  "@type": "croissant:OutputSpec",
  "@id":   "<base>#outputSpec",
  "croissant:schema": {
    "@type": "croissant:RecordSet",
    "croissant:field": [
      { "@type": "croissant:Field", "schema:name": "score", "croissant:dataType": "xsd:float" }
    ]
  }
}
```

**(b) Free-form string** - generated text:
```json
{
  "@type": "croissant:Field",
  "schema:name": "generated_text",
  "croissant:dataType": "xsd:string"
}
```

**(c) Vector / array** - embeddings or logits:
```json
{
  "@type": "croissant:Field",
  "schema:name": "embedding",
  "croissant:dataType": "xsd:float",
  "croissant:repeated": true
}
```

**(d) Categorical** - constrained label (e.g. MMLU's A/B/C/D):
```json
{
  "@type": "croissant:Field",
  "schema:name": "answer",
  "croissant:dataType": "xsd:string",
  "schema:valuePattern": "^[A-D]$",
  "schema:description": "The predicted answer choice (A, B, C, or D)."
}
```

### Evaluation block

```json
"croissant:evaluation": {
  "@type": "croissant:EvaluationSpec",
  "@id":   "<base>#evaluationSpec",
  "croissant:expectedMetric": ["Accuracy", "F1-Score"]
}
```

`expectedMetric` values are strings (or IRIs). Pull metric names directly from the paper's wording.

### Subtasks (if the benchmark has sub-categories)

For each sub-category:

```json
{
  "@type": "croissant:TaskProblem",
  "@id":   "<base>#<sub_slug>",
  "schema:name":        "Benchmark - Sub-category",
  "schema:description": "...",
  "croissant:input":  { /* same dataset OR pre-processed slice */ },
  "croissant:output": { "@id": "<base>#outputSpec" },
  "croissant:evaluation": { "@id": "<base>#evaluationSpec" }
}
```

**Deduplication and Referencing**: If a `subTask` uses the same field (like `input`, `output`, or `evaluation`) that already appears in the parent task or another subtask, you must still include that field in the subtask. However, instead of repeating the full object, use the `@id` property to point to the existing definition. For example: `"croissant:input": { "@id": "parent_input_id" }`. This ensures constraints are correctly checked while avoiding duplication.

### Reference Examples

Check the `tasks/benchmark_examples/` directory in the repository for examples of Croissant Tasks files (such as MMLU). These examples show how to structure `TaskProblem` and `TaskSolution` files for different types of benchmarks, including usage of subtasks and deduplication. You can fetch them on demand from the repository.

---

## Step 5: Build `TaskSolution` JSON-LDs (one per reported baseline)

For each model the paper reports results for, write `{{results_dir}}/solutions/<model_slug>.jsonld`.

Skip this step if the paper reports no baselines.

### Required structure

```json
{
  "@context": { /* same @context as problem */ },
  "@type": "croissant:TaskSolution",
  "@id":   "<base>_solution_<model_slug>",
  "schema:name": "<Benchmark> Solution - <Model Name>",

  "schema:isBasedOn": { "@id": "<base>" },

  "croissant:implementation": {
    "@type": "schema:SoftwareApplication",
    "@id":   "<base>_solution_<model_slug>#implementation",
    "schema:name": "<Model Provider / API>"
  },

  "croissant:execution": {
    "@type": "croissant:ExecutionConfig",
    "@id":   "<base>_solution_<model_slug>#execution",
    "croissant:hyperparameter": [
      { "@type": "schema:PropertyValue", "schema:name": "<hyperparameter_name>", "schema:value": "<value>" }
      /* List hyperparameters determined from the paper (e.g., temperature, few-shot k, etc.) */
    ]
  },

  "croissant:output": {
    "@type": "schema:Dataset",
    "@id":   "urn:uuid:<random-or-paper-ref>",
    "schema:name": "<Model>'s outputs on <Benchmark>"
  },

  "croissant:subTask": [ /* one TaskSolution per subtask, if applicable */ ],

  "croissant:evaluation": {
    "@type": "croissant:EvaluationTask",
    "@id":   "<base>_evaluation_<model_slug>",
    "schema:name": "Evaluation of <Model> on <Benchmark>",
    "schema:isBasedOn":              { "@id": "<base>" },
    "croissant:evaluatedSolution":   { "@id": "<base>_solution_<model_slug>" },
    "croissant:evaluationResults": [
      { "@type": "croissant:EvaluationResult",
        "croissant:metric": "<Metric Name>",
        "croissant:value":  "<Reported Value or Dynamic Result>",
        "schema:description": "<Optional: e.g., 'Overall Average Accuracy' or 'Logged from run'>" }
      /* Populate with results reported in the paper or left to be logged by the implementation */
    ]
  }
}
```

### SHACL constraints you MUST satisfy (TaskSolutionShape)

- **MUST** have `schema:isBasedOn` --> `TaskProblem` or an IRI.
- **MUST NOT** have any `InputSpec`, `OutputSpec`, `ImplementationSpec`, or `EvaluationSpec` anywhere as direct values of `input`, `output`, `implementation`, or `evaluation`. Solutions are concrete - use `schema:Dataset`, `schema:SoftwareApplication`, `EvaluationTask`, etc.
- **MUST** have EITHER a concrete `implementation` (not `ImplementationSpec`), OR have `subTask`s where every subtask has a concrete implementation.
- **EvaluationResult** requires both `croissant:metric` (string or URL) and `croissant:value` (number, string, or QuantitativeValue).
- **EvaluationTask** requires exactly one `croissant:evaluatedSolution` pointing at a `TaskSolution`.

### Tips

- Metric values can be JSON numbers (`25.9`) or strings (`"25.9"`). Match what feels natural but keep it consistent across all results within a solution.
- When the paper reports sub-category accuracies, mirror the problem's subtask structure in the solution's `subTask` array - each sub-solution references the matching sub-problem via `schema:isBasedOn` and has its own `EvaluationTask`. See `mmlu_solution_small_fewshot.jsonld` for the canonical pattern.
- Reference the parent solution's `ExecutionConfig` / `SoftwareApplication` in sub-solutions by `@id` - don't redefine hyperparameters per subtask.
- `croissant:output` on a solution is a concrete `schema:Dataset` (the outputs it produced). Use `urn:uuid:<something>` as the `@id` if no real URL exists.
- Model slug: kebab-case, lowercased, no spaces (e.g. `gpt4-fewshot`, `llama2-70b-zeroshot`).

---

## Step 6: Evaluate Outputs (programmatic)

Use the Python validator available in the latest commit of PR #1017 (e.g., `validator.py`) to check constraints in the files generated. This validator is the official way to verify that your generated JSON-LD files conform to the Croissant Tasks specification and shapes.

Run the validator on each generated file:
```bash
# Example usage (adjust path to validator.py as needed)
python3 path/to/validator.py {{results_dir}}/problem.jsonld
python3 path/to/validator.py {{results_dir}}/solutions/<slug>.jsonld
```

Capture the output of the validator to identify any non-conformance issues.

Per-file status mapping:

| Status | Criteria |
|--------|----------|
| `PASS` | JSON parses AND SHACL conforms |
| `FAIL` | JSON parse error OR SHACL non-conformance |

---

## Step 7: Iterate on Errors (max 3 rounds)

If any file shows `FAIL`:

1. Read the SHACL `report` text for that file - it identifies the failing shape, the focus node, and the constraint.
2. Apply the targeted fix from the table below, or reason from the shape message directly.
3. Rewrite the offending `.jsonld` file.
4. Re-run Step 6.
5. Stop at 3 iterations or when everything passes.

### Common fixes

| SHACL message | Fix |
|---------------|-----|
| `A TaskProblem must have at least one property ... that is a spec class` | Add an `OutputSpec` (most common) with a `RecordSet` + at least one `Field` with `dataType`. |
| `croissant:output must point to a Dataset, SoftwareSourceCode, or OutputSpec` | On a Solution, output must be `schema:Dataset` (not `OutputSpec`). On a Problem, it's usually `OutputSpec`. |
| `croissant:evaluation must be an EvaluationTask` (on a TaskShape / non-Problem) | Use `EvaluationTask` (concrete) on Solutions; `EvaluationSpec` is only allowed on Problems. |
| `A TaskSolution must be formally linked to a TaskProblem via schema:isBasedOn` | Add `"schema:isBasedOn": { "@id": "<problem_@id>" }`. |
| `A TaskSolution cannot have an OutputSpec/InputSpec/ImplementationSpec/EvaluationSpec` | Replace the offending Spec with its concrete counterpart. |
| `TaskSolution must have at least one concrete implementation ...` | Add `croissant:implementation` as `schema:SoftwareApplication` or `schema:SoftwareSourceCode`. If using subTasks, ensure every subTask has concrete implementation. |
| `croissant:metric/value is required` (on EvaluationResult) | Every `EvaluationResult` needs both `croissant:metric` and `croissant:value`. |
| `croissant:evaluatedSolution must point to exactly one TaskSolution` | On every `EvaluationTask`, set exactly one `{ "@id": "<solution_@id>" }`. |
| `croissant:schema must point to a RecordSet` | On `InputSpec` / `OutputSpec`, make sure `croissant:schema` resolves to a node typed `croissant:RecordSet`. |
| `A RecordSet must have at least one field` / `A Field must have a dataType` | Every RecordSet needs >=1 Field; every Field needs `croissant:dataType` as an IRI (e.g. `"xsd:string"`, `"xsd:float"`). |
| JSON-LD parse error | Check for trailing commas, unquoted keys, missing `@context`. |
| RDF graph is empty for a file | Ensure `@type` uses the `croissant:` or `schema:` prefix (not a bare identifier). |

After 3 iterations, keep the best attempt and flag every remaining failure in `summary.md`.

---

## Step 8: Write Executive Summary

Write `{{results_dir}}/summary.md`:

```markdown
# Croissant Tasks Report: <Benchmark Name>

## Overview
- **Date**: <run date>
- **Paper**: <title, authors>
- **PDF**: {{pdf_location}}
- **Paper URL**: {{paper_url}}
- **Dataset URL**: {{dataset_url}}
- **@id base**: <what you chose>

## Files emitted
| File | Type | Size |
|------|------|------|
| problem.jsonld | TaskProblem | ... bytes |
| solutions/<model>.jsonld | TaskSolution | ... bytes |

## TaskProblem extraction

### High-confidence fields (explicitly stated)
| Field | Value | Paper section |
|-------|-------|---------------|
| name | ... | ... |
| description | ... | Abstract |
| input dataset | ... | ... |
| output schema | ... | ... |
| metrics | ... | ... |
| subtasks | ... | ... |

### Inferred fields (medium confidence)
| Field | Value | Rationale |
|-------|-------|-----------|
| ... | ... | ... |

### Skipped (paper was silent)
| Field | Reason |
|-------|--------|
| ... | Not mentioned |

## Solutions extracted

| Model | Hyperparameters | Metrics --> Values | Subtask breakdown |
|-------|-----------------|------------------|-------------------|
| ... | T=0.0, top-p=1.0 | Accuracy=..., F1=... | humanities/STEM/... |

## Validation results

| File | JSON | SHACL | Iterations | Remaining errors |
|------|------|-------|-----------|------------------|
| problem.jsonld | PASS | PASS | 1 | - |
| solutions/gpt4.jsonld | PASS | PASS | 2 | - |

## Limitations / caveats
- <what you couldn't determine>
- <things a human should manually verify>
```

---

## Step 9: Write Validation Report

Write `{{results_dir}}/validation_report.json`:

```json
{
  "version": "1.0.0",
  "run_date": "<ISO8601>",
  "parameters": {
    "pdf_location":  "{{pdf_location}}",
    "paper_url":     "{{paper_url}}",
    "dataset_url":   "{{dataset_url}}"
  },
  "stages": [
    { "name": "setup",               "passed": true, "message": "Shapes + ontology downloaded, deps installed" },
    { "name": "paper_analysis",      "passed": true, "message": "Paper read, task + N baselines extracted" },
    { "name": "problem_generated",   "passed": true, "message": "problem.jsonld written" },
    { "name": "solutions_generated", "passed": true, "message": "N solution files written" },
    { "name": "json_validity",       "passed": true, "message": "All files parse as JSON" },
    { "name": "shacl_conformance",   "passed": true, "message": "All files conform to Croissant Tasks SHACL shapes" },
    { "name": "report_generation",   "passed": true, "message": "summary.md + validation_report.json written" }
  ],
  "per_file": [
    { "file": "problem.jsonld",               "json_valid": true, "shacl_conforms": true, "iterations": 1 },
    { "file": "solutions/<model>.jsonld",     "json_valid": true, "shacl_conforms": true, "iterations": 1 }
  ],
  "results":        { "pass": 0, "fail": 0 },
  "overall_passed": true,
  "iterations":     1,
  "output_files": [
    "{{results_dir}}/problem.jsonld",
    "{{results_dir}}/solutions/<model>.jsonld",
    "{{results_dir}}/summary.md",
    "{{results_dir}}/validation_report.json"
  ]
}
```

Fill `results.pass` / `results.fail` with actual per-file counts.

---

## Step 10: Final Checklist (MANDATORY - do not skip)

### Verification Script

```bash
echo "=== FINAL OUTPUT VERIFICATION ==="
R="{{results_dir}}"

# Required files
for f in "$R/problem.jsonld" "$R/summary.md" "$R/validation_report.json"; do
  if [ ! -s "$f" ]; then
    echo "FAIL: $f missing or empty"
  else
    echo "PASS: $f ($(wc -c < "$f") bytes)"
  fi
done

# Solutions (0+ files allowed, but if any exist they must be non-empty)
if [ -d "$R/solutions" ]; then
  count=$(find "$R/solutions" -name "*.jsonld" -size +0 2>/dev/null | wc -l | tr -d ' ')
  echo "INFO: $count solution file(s)"
fi

# JSON validity
python3 -c "import json,pathlib; [json.loads(p.read_text()) for p in pathlib.Path('$R').rglob('*.jsonld')]" \
  && echo "PASS: all .jsonld files parse" \
  || echo "FAIL: at least one .jsonld does not parse"

python3 -c "import json; d=json.load(open('$R/validation_report.json')); assert 'overall_passed' in d" \
  && echo "PASS: validation_report.json has overall_passed" \
  || echo "FAIL: validation_report.json malformed"

# SHACL re-check using official validator
echo "INFO: Running official validator..."
for f in "$R/problem.jsonld" $(find "$R/solutions" -name "*.jsonld" 2>/dev/null); do
  if [ -f "$f" ]; then
    python3 path/to/validator.py "$f" \
      && echo "PASS: $f conforms" \
      || echo "FAIL: $f does not conform"
  fi
done
```

### Checklist

- [ ] `problem.jsonld` exists, parses, `@type` is `croissant:TaskProblem`, has at least one Spec (usually `OutputSpec`).
- [ ] `solutions/` dir exists. Every file (if any) has `@type` `croissant:TaskSolution`, `schema:isBasedOn` pointing at the problem, and a concrete `implementation`.
- [ ] `summary.md` follows the Step 8 template and documents extraction confidence.
- [ ] `validation_report.json` follows the Step 9 schema and `overall_passed` reflects reality.
- [ ] The verification script printed PASS for every required line.

**If ANY item fails, go back and fix it. Do NOT finish until all items pass.**

---

## Tips

- **The paper is the primary source of truth.** Don't fabricate metrics, hyperparameters, or subtask structure. If a value isn't in the paper, leave it out and document the gap in `summary.md`.
- **@id values must be globally unique within a document.** Use fragment identifiers (`#outputSpec`, `#execution`, `#humanities_sol`) to namespace them under your base IRI.
- **Use `@id` deduplication.** Whenever the same `OutputSpec` / `EvaluationSpec` / `ExecutionConfig` applies across subtasks, define it once at the top with a named `@id` and reference it by `{ "@id": "..." }` everywhere else. This keeps the JSON small and makes the RDF graph coherent.
- **Controlled vocabulary is the paper's vocabulary.** The Tasks spec itself has no fixed enum for `expectedMetric` or hyperparameter names - copy the wording the paper uses ("Accuracy" vs "accuracy", "Exact Match" vs "EM").
- **Multiple inputs.** `croissant:input` can be an array - useful for few-shot tasks (test data + exemplars). See MMLU's `*_fewshot` subtasks.
- **`xsd:` datatype IRIs must be written as strings** (`"xsd:string"`, not `{ "@id": "xsd:string" }`). The SHACL shape uses `sh:nodeKind sh:IRI`, and the standard prefix expansion handles both forms, but keep it as a plain string for readability.
- **Local Spec Files.** If operating within the Croissant Tasks repository, use the local copies of `croissant-tasks.ttl` and `croissant-tasks-shapes.ttl` for validation instead of downloading them from GitHub.
- **Mapping Tables to Solutions.** Each row in a main results table of a paper typically corresponds to a `TaskSolution`. The row represents a specific model/baseline, and the columns usually represent different subtasks or metrics.
- **Pretty-print JSON-LD** with 2-space indent - these files are read by humans.
- **When in doubt, mirror the MMLU example.** It's the canonical worked example from PR #1017 and covers every shape (problem + subtasks + solution + evaluation + dedup).  
\end{lstlisting}

We also propose the following standard prompt for users to pass the instruction to the agent.
\begin{lstlisting}[basicstyle=\ttfamily\scriptsize, frame=single, breaklines=true]
# Request: Convert ML Benchmark Paper to Croissant Tasks

Please help me convert a scientific paper describing a machine learning benchmark into the **MLCommons Croissant Tasks** format.

You MUST use the pdf2ct skill file to complete this task, as it contains the specific ontology mapping rules and validation steps required for this project.

Here are the details for the paper I want to convert:

## USER INPUT: Paper Details
  **PDF Location**: [Insert local path OR URL to the PDF file here]
  **Dataset URL (Optional)**: [Insert link to the dataset if applicable, or leave blank]
  **Baselines to Extract (Optional)**: [List specific models/baselines to extract (e.g., "GPT-4, Claude 3"), or leave blank to ONLY generate the TaskProblem definition.]

## What I Expect
1.  Follow the steps in the pdf2ct skill file carefully.
2.  Generate the problem.jsonld file.
3.  **If baselines were specified above**, also generate the corresponding solutions/*.jsonld files for those specific models. If the field was left blank, do not generate any solution files.
4.  Validate all generated files using the official validator as specified in the skill.
5.  Provide the required summary.md and validation_report.json files in the results directory.

Please confirm that you can access the PDF and let me know if you need any clarification before starting! 
\end{lstlisting}

\section{Skill File and Prompt for Agentic Code Generation}
\label{app:skills-ct2code}
We propose the following \texttt{SKILL.md} file to assist the agent to transform a Croissant Tasks specification into a concrete implementation.
\begin{lstlisting}[basicstyle=\ttfamily\scriptsize, frame=single, breaklines=true]
---
name: ct2code
description: >-
  Guides an agent to go from Croissant Tasks TaskProblem files and a baseline description to actual code implementations of the benchmark and a particular baseline, and generating the corresponding TaskSolution file.
---

# Croissant Tasks to Code - Agent Runbook

## Objective

You are given a Croissant Tasks **`TaskProblem`** file (JSON-LD) describing a benchmark and a description of a **baseline** (model, hyperparameters, prompt templates) that you want to test on it. Your job is to generate the code implementing that solution and the corresponding Croissant Tasks **`TaskSolution`** file.

The generated code should perform the following tasks:
1.  **Data loading**: Download and read the input dataset specified in the problem file.
2.  **Sub-task structure**: Implement necessary data processing and structure the code taking into account the `subTask` structure of the benchmark.
3.  **Implementation of baseline**: Implement the model API call (or local execution) for each sub-task, using the correct hyperparameters and prompts.
4.  **Evaluation**: Load or implement the evaluation metrics, compute them by comparing the output with golden labels, and log the metrics.
5.  **Croissant Task file writing**: Write the `TaskSolution` file, pointing to outputs and execution info, and correctly linking to the problem.

---

## Guidance

### Step 1: Analyze the `TaskProblem` File

Read the `TaskProblem` file to understand the benchmark structure:
- **Inputs**: Look for `croissant:input`. It might point to a Hugging Face dataset or a specific file.
- **Outputs**: Look for `croissant:output` and `OutputSpec` to understand the expected format of the predictions.
- **Subtasks**: Check if there are `croissant:subTask`s. If so, the implementation should likely handle them separately or iterate over them.
- **Evaluation**: Look for `croissant:evaluation` and `expectedMetric` to know what to measure.

### Step 2: Gather Baseline Information

You will need additional information not present in the `TaskProblem` file (which only defines the problem):
- **Model**: Which model to use (e.g., Gemini, GPT).
- **Hyperparameters**: Temperature, max tokens, etc.
- **Prompts**: Prompt templates for each sub-task if applicable.

This information can be provided as a **natural language description** in the user request or extracted from a referenced source (e.g., a paper's appendix). Alternatively, it can be provided as a **"skeleton" Croissant Tasks `TaskSolution` file** that contains the configuration but lacks the final results. The implementation should read this configuration and overwrite the file with the actual results after execution.

### Step 3: Generate Implementation Code

Generate Python code (or the language requested) that does the following:

#### 3.1 Data Loading
- Use appropriate libraries (e.g., `datasets` for Hugging Face) to load the data.
- Handle any authentication or downloading steps needed.

#### 3.2 Processing and Sub-tasks
- If the task has sub-tasks, structure the code to loop over them or call specific handlers.
- Align the data loading with the specific cuts/splits defined for each sub-task.

#### 3.3 Baseline Execution
- Implement the API call or model inference.
- Ensure hyperparameters are correctly passed.
- Apply prompt templates to the input data.

#### 3.4 Evaluation
- Extract golden labels from the loaded dataset.
- Compare model outputs with golden labels.
- **Implement all the metrics specified in the `TaskProblem`**. Do not use proxies or placeholders unless explicitly allowed by the user or strictly restricted by the environment.
- **Identify needed libraries**: If specific external libraries are required for proper metric calculation, the agent should check if they are available in the environment. If not, **ask the user to install them** instead of silently falling back to simplified proxies or placeholders.
- **Data Modality Checks**: Perform common-sense checks based on the data modality:
    - **Vision**:
        - **Coordinate Space**: Ensure prediction coordinates match ground truth scale (e.g., pixels vs normalized).
        - **Image Processing**: Verify color channel order (RGB vs BGR) and that resizing doesn't break coordinate mappings.
    - **Language**:
        - **Normalization**: Ensure consistent handling of casing, punctuation, and whitespace between predictions and references.
        - **Tokenization**: Be aware of different tokenization strategies that might affect metrics like BLEU or METEOR.
- **Robust Parsing**: Anticipate that LLMs may not follow strict output formats (like JSON) or might get truncated. Implement robust parsing with fallbacks or simpler plain text formats if needed.

#### 3.5 Output and Solution Generation
- Save the raw outputs (predictions) to a file.
- Save execution metadata (hyperparameters, timestamps) to a file.
- Generate the `TaskSolution` JSON-LD file.
- **Incremental Execution**: For long-running evaluations, consider implementing an incremental execution pattern by saving intermediate results to a file and skipping already processed samples on restart.

### Step 4: Generate `TaskSolution` File

The code should output a `TaskSolution` file that conforms to the Croissant Tasks spec.
- Set `@type` to `croissant:TaskSolution`.
- Link to the problem via `"schema:isBasedOn": { "@id": "<problem_id>" }`.
- Fill in `croissant:execution` with the actual hyperparameters used.
- **Correctly point to the location of the generated outputs and implementation files** (e.g., using relative paths within the repository or universally accessible URIs).
- Fill in `croissant:evaluation` with an `EvaluationTask` containing `EvaluationResult`s for each metric with their **latest concrete values**.
- **The `TaskSolution` MUST respect the `subTask` structure of the `TaskProblem`.** If the problem file defines `croissant:subTask`s, the solution file must also contain a `croissant:subTask` list, with each element being a `TaskSolution` pointing to the specific sub-task ID and containing its specific implementation, output, and evaluation results.

---

## Verification Plan for the Generated Code

The generated code should be verifiable. The agent should include instructions or tests to:
1.  **Dry Run**: Run on a small subset of data to verify the pipeline works.
2.  **Schema Check**: Verify that the generated `TaskSolution` file passes the Croissant Tasks validator.
3.  **Metric Verification**: Verify that the computed metrics match expected values if a small test case with known results is available.
4.  **Metric Plausibility Check**: If metrics are unexpectedly low (e.g., zero or near zero), double-check the implementation against the metric description in the paper or problem file. Common issues include coordinate format mismatches or parsing failures of LLM outputs.

\end{lstlisting}

We also propose the following standard prompt for users to pass the instruction to the agent.
\begin{lstlisting}[basicstyle=\ttfamily\scriptsize, frame=single, breaklines=true]
I want to evaluate a baseline on a benchmark defined in a Croissant Tasks file. Please use the ct2code skill to generate the implementation code and the corresponding TaskSolution file.

## 1. Task Problem
**File Path**: [Path to the TaskProblem JSON-LD file, e.g., problem.jsonld]

## 2. Baseline Specification
(Choose one of the options below)

### Option A: Natural Language Description
**Model**: [e.g., Gemini 1.5 Flash]
**Hyperparameters**:
  - [e.g., temperature: 0.0]
  - [e.g., max_output_length: 1024]
**Prompt Templates**:
  - **Sub-task [Sub-task Name/ID]**: [Prompt template with placeholders like {input}]
  - **Sub-task [Another Sub-task]**: [Prompt template]

### Option B: Skeleton TaskSolution File
**File Path**: [Path to the skeleton JSON-LD file containing config but no results]

## 3. Environment & Authentication
**API Keys**: [e.g., Use GEMINI_API_KEY environment variable]
**Other requirements**: [e.g., Needs access to specific internal services]

## 4. Outputs
**Generated Code Destination**: [Path to save the generated Python script]
**Raw Outputs Destination**: [Path to save model predictions]
**TaskSolution Destination**: [Path to save the final TaskSolution JSON-LD]

## 5. Additional Context
[e.g., The dataset is on Hugging Face at 'user/dataset']
[e.g., Use standard accuracy for evaluation]
\end{lstlisting}

\section{Prompt for Baseline}
The baseline we compare against consists of generating implementations directly from the paper PDF. Here is the prompt employed for that.

\begin{lstlisting}[basicstyle=\ttfamily\scriptsize, frame=single, breaklines=true]
You will be given a link to the PDF of a benchmark paper in machine learning.
Your task is to generate an implementation allowing the reproduction of a baseline of the paper.
DO NOT launch the full evaluation without permission.
You may launch test calls to the API to debug your code.
Implement mechanisms to save your progress during full evaluation, so that you can continue from where you stopped in case of errors, and implement relaunching mechanisms in case of 503 Server unavailable errors.

## 1. Problem
**Paper PDF URL or path**: [URL or path to paper]

## 2. Baseline Specification
**Baseline / model to test**: [Name of the baseline or model]
**Hyperparameters**:
  - [e.g. temperature: 0.1]
**Prompt Templates**:
  - [Path to or direct writing of prompt templates]

## 3. Environment & Authentication
**API Keys**: [e.g. Use GEMINI_API_KEY environment variable]

## 4. Outputs
**Generated Code Destination**: [Where to save generated implementation]
**Raw Outputs Destination**: [Where to save generated outputs]
\end{lstlisting}

\section{Details on Experiments}
\label{app:details_exp}
For each benchmark, we report the differences between results described in the paper and those obtained by running our agentic implementations. Notice that in some cases we do not run the benchmark on the entire dataset, for reasons of costs.

\subsection{Details for AbsenceBench Experiments}

For AbsenceBench~\citep{fu2026absence}, we run full-scope reproduction on the full validation split with 3,278 examples under two \textbf{prompt-template settings}.
In \texttt{paper\_only}, prompts are instantiated from the paper's Appendix A templates.
In \texttt{ct\_only}, prompts are instantiated from Croissant Task semantics.
In both settings, inference uses the same dataset fields (\texttt{original\_context}, \texttt{modified\_context}).
As in the other detailed benchmark sections, we decompose implementation into three components:
\begin{itemize}
  \item[\textbf{C1}] \textbf{Data loading.} Load all three task configs from \href{https://huggingface.co/datasets/harveyfin/AbsenceBench}{\texttt{harveyfin/AbsenceBench}}: \texttt{poetry}, \texttt{numerical}, and \texttt{github\_prs}. Preserve per-instance IDs for deterministic merging and verification.
  \item[\textbf{C2}] \textbf{Generation model configuration.} Call Claude~4~Sonnet with \texttt{temperature}=0.0 and \texttt{max\_tokens}=4096. For ablation integrity, workers are condition-isolated, so \texttt{paper\_only} and \texttt{ct\_only} never read each other's prompt or response files.
  \item[\textbf{C3}] \textbf{Evaluation pipeline.} Parse omission predictions and compute domain-level F1 and Exact Match. Then compute bootstrap confidence intervals with 1,000 resamples from saved raw outputs.
\end{itemize}

The first full-scope run revealed instruction-collapse failures on long-context prompts. We therefore reran with chunked execution at 80 rows per chunk, explicit row-count and ID-order QA gates, and targeted chunk reruns before finalization. Table~\ref{tab:absencebench_metric_comparison} reports side-by-side results. To align with the paper's headline reporting, $\Delta$ is computed against the paper's Claude 3.7 Sonnet \textit{thinking} baseline, marked with $^\star$ in the table.

\begin{table}[t]
\centering
\setlength{\tabcolsep}{5pt}
\renewcommand{\arraystretch}{1.15}
\resizebox{\linewidth}{!}{
\begin{tabular}{llccccc}
\toprule
\textbf{Model} & \textbf{Metric}
& \textbf{Reported}
& \multicolumn{2}{c}{\textbf{PDF$\rightarrow$CT$\rightarrow$code}}
& \multicolumn{2}{c}{\textbf{PDF$\rightarrow$code (direct)}} \\
\cmidrule(lr){4-5} \cmidrule(lr){6-7}
& & \textbf{(\%)} & \textbf{Obtained [95\% CI]} & \textbf{$\Delta$}
& \textbf{Obtained [95\% CI]} & \textbf{$\Delta$} \\
\midrule
\multirow{4}{*}{Claude 4 Sonnet}
& Poetry Micro-F1      & 72.70$^\star$ & 98.62 [98.47, 98.76] & +25.92 & 98.63 [98.36, 98.87] & +25.93 \\
& Numerical Micro-F1   & 96.00$^\star$ & 100.00 [100.00, 100.00] & +4.00 & 100.00 [100.00, 100.00] & +4.00 \\
& GitHub PRs Micro-F1  & 40.00$^\star$ & 23.36 [21.51, 25.16] & -16.64 & 44.65 [41.79, 47.36] & +4.65 \\
& Mean Domain Micro-F1 & 69.60$^\star$ & 73.99 [73.38, 74.61] & +4.39 & 81.09 [80.20, 82.00] & +11.49 \\
\midrule
\multirow{2}{*}{Supplementary}
& Pooled micro-F1 & -- & 91.58 [90.72, 92.37] & -- & 96.06 [95.57, 96.50] & -- \\
& Exact Match     & -- & 61.41 [59.58, 63.15] & -- & 63.30 [61.41, 65.10] & -- \\
\bottomrule
\end{tabular}
}
\vspace{1em}
\caption{Comparison between reported AbsenceBench results and our reproduced results under two pipelines. In this benchmark, \texttt{PDF$\rightarrow$CT$\rightarrow$code} corresponds to \texttt{ct\_only}, and \texttt{PDF$\rightarrow$code (direct)} corresponds to \texttt{paper\_only}. Mean Domain Micro-F1 is the unweighted mean across the three domain-level Micro-F1 scores. For obtained metrics, 95\% confidence intervals are reported in brackets and computed via bootstrapping with $N{=}1000$ iterations. $\Delta$ denotes obtained minus reported. \textbf{Important note:} values marked with $^\star$ are paper-reported results measured with Claude 3.7 Sonnet (thinking), which is now discontinued; they are included as a historical reference for comparability.}
\label{tab:absencebench_metric_comparison}
\vspace{-1em}
\end{table}

Run-level progression on the same full validation scope is as follows. The initial ungated full run produced pooled micro-F1 of 71.26 for \texttt{paper\_only} and 58.61 for \texttt{ct\_only}. After chunking and QA gates, the final run reached 96.06 and 91.58. Exact Match moved from 36.61 to 63.30 for \texttt{paper\_only} and from 38.90 to 61.41 for \texttt{ct\_only}.

Most of the difference between the two pipelines comes from \texttt{github\_prs}. On \texttt{poetry}, both pipelines are almost identical (98.62 vs.\ 98.63 Micro-F1), and on \texttt{numerical}, both are perfect (100.00 Micro-F1). The large gap appears in \texttt{github\_prs}: \texttt{PDF$\rightarrow$CT$\rightarrow$code} gets 23.36 Micro-F1, while \texttt{PDF$\rightarrow$code (direct)} gets 44.65 Micro-F1. The CT-only pipeline returned longer candidate lists. Across 887 examples, CT-only predicts 26.12 lines per example on average, while the direct pipeline predicts 13.10 lines, and the gold average is 3.77. This creates many extra lines: CT-only has 20,071 false positives versus 8,277 for the direct pipeline. Both pipelines still recover most true omissions, so recall remains high (92.61 vs.\ 99.91), but CT-only is noisier, which lowers precision (13.36 vs.\ 28.75) and therefore Micro-F1. This pattern is consistent with the prompt sources: the direct pipeline uses the paper's domain-specific Appendix A template, which explicitly narrows attention to changed diff lines, while the CT-only pipeline uses a more generic omission prompt.

For context, the paper also reports a no-thinking Claude 3.7 Sonnet baseline with Poetry Micro-F1 of 73.5, Numerical Micro-F1 of 91.4, GitHub PRs Micro-F1 of 35.7, and mean domain Micro-F1 of 66.9. We do not report a single fixed $k$ trajectory for this benchmark because the final run used a scripted QA loop with targeted chunk reruns instead of a capped interaction protocol.

\subsection{Details for CoRe Experiments}

We restrict our reproduction to the \textbf{Control-Dependency / Trace subtask} ($n{=}489$) of CoRe Lite, which spans 1,584 items across all subtasks, due to API credit constraints. We decompose the reproduction into three implementation components:
\begin{itemize}
  \item[\textbf{C1}] \textbf{Data loading.} Loading the \href{https://huggingface.co/datasets/lt-asset/CoRe}{dataset from HuggingFace} and filtering to the Control-Dependency / Trace subtask.
  \item[\textbf{C2}] \textbf{Generation model configuration.} Calling Gemini~2.5~Pro; the dataset's \texttt{prompt} field is self-contained, with instructions, definitions, output format, and 5--7 few-shot examples pre-baked per row, so no prompt construction is required from the agent. The only model-side hyperparameter of note is \texttt{max\_output\_tokens}.
  \item[\textbf{C3}] \textbf{Evaluation pipeline.} Parsing the structured trace outputs produced by the model and computing the paper's two metrics: F1~Score and Correct~Trace~Rate.
\end{itemize}

The values of the metrics reported in the paper and obtained by our implementation are shown in Table~\ref{tab:core_metric_comparison}.

\begin{table*}[htbp]
\centering
\setlength{\tabcolsep}{5pt}
\renewcommand{\arraystretch}{1.15}
\resizebox{\linewidth}{!}{
\begin{tabular}{llccccc}
\toprule
\textbf{Model} & \textbf{Metric}
& \textbf{Reported}
& \multicolumn{2}{c}{\textbf{PDF$\rightarrow$CT$\rightarrow$code}}
& \multicolumn{2}{c}{\textbf{PDF$\rightarrow$code (direct)}} \\
\cmidrule(lr){4-5} \cmidrule(lr){6-7}
& & \textbf{(\%)} & \textbf{Obtained [95\% CI]} & \textbf{$\Delta$}
& \textbf{Obtained [95\% CI]} & \textbf{$\Delta$} \\
\midrule
\multirow{2}{*}{Gemini 2.5 Pro}
& F1 Score                            & 92.49 & 94.58 [92.37, 96.46]    & +2.09 & 80.20 [75.75, 84.11] & -12.29 \\
& Correct Trace Rate     & 92.26 & 100.00 [100.00, 100.00] & +7.74 & 96.93 [93.96, 99.37] & +4.67  \\
\bottomrule
\end{tabular}
}
\caption{Comparison between CoRe paper Table~4 results (Gemini~2.5~Pro, evaluated on the full CoRe Lite, $n{=}1{,}584$) and our two reproductions on the Control-Dependency / Trace subtask only ($n{=}489$ for each pipeline). Obtained values are point estimates with 95\% bootstrap confidence intervals.}
\label{tab:core_metric_comparison}
\end{table*}

In the {PDF$\rightarrow$code (direct) pipeline}, the agent produced a correct implementation under zero guiding prompts. We subsequently issued two prompts: one restricting execution to the Control-Dependency / Trace subtask due to credit constraints, and one enabling parallel API calls for faster iteration. Both were scope and efficiency adjustments; the original sequential, all-subtasks implementation was already correct.

In the {PDF$\rightarrow$CT$\rightarrow$code (no-PDF) pipeline} in the \texttt{pdf2ct} stage, three hyperparameters from CoRe paper's Appendix~E, the decoding \texttt{temperature}, the \texttt{max\_output\_tokens=2048} cap, and the up-to-three fallback-retry count, were missed; the coding stage proceeded with this incomplete skeleton. A later re-run of \texttt{pdf2ct} on the same paper did capture all three, confirming the omission was an extraction inconsistency rather than a schema limitation: the schema carries the slots, and \texttt{pdf2ct} can fill them. The likely cause is long-context handling; all three details appear within four sentences on page~30 of the 55-page paper, and extraction reliability becomes sensitive to how the agent paces its reading when a critical specification is buried deep in a long document.

For the coding stage of the {PDF$\rightarrow$CT$\rightarrow$code pipeline}, the agent had access only to the resulting skeleton files and the HuggingFace dataset; the paper PDF was withheld. Under the same scope restriction as the direct run, it produced an end-to-end implementation covering data loading, sub-task structuring, model invocation, parsing, metric computation, and \texttt{TaskSolution} regeneration. Structural correctness was preserved because the dataset's \texttt{prompt} field is self-contained: instructions, definitions, output format, and few-shot examples are pre-baked per row, so the model requires nothing from the paper to be prompted correctly. The agent did not implement Appendix~E's fallback-retry mechanism, as that specification was not present in the CT skeleton; on this subtask the omission did not affect coverage, as 0 out of 489 responses were invalid. As with the direct run, one post-hoc guiding prompt enabled parallel API calls; the original sequential implementation was already working correctly.

\paragraph{Invalid-response gap.}
The \texttt{PDF$\rightarrow$code} (direct) run produced 15 invalid responses out of 489, whereas the \texttt{PDF$\rightarrow$CT$\rightarrow$code} (no-PDF) run produced 0. Both pipelines send identical prompt text, the \texttt{prompt} field of the HuggingFace \texttt{lt-asset/CoRe} dataset, which already contains the paper's base instructions and 5--7 pre-rendered few-shot examples. The most likely mechanism behind the gap is the output-token limit. Gemini~2.5~Pro is a reasoning model whose hidden reasoning tokens count against the output budget; the \texttt{max\_output\_tokens=2048} setting specified in Appendix~E can therefore exhaust the budget before any visible output is produced. The direct agent read this setting from the paper and applied it; the no-PDF agent, lacking the hyperparameter, used the API default and left sufficient headroom for both reasoning and structured output. The F1 gap between the two pipelines ($-12.29$ pp) is largely attributable to these 15 invalid responses rather than to a fundamental implementation difference.

\paragraph{Snapshot note.}
The paper evaluates Gemini~2.5~Pro using the pinned snapshot \texttt{gemini-2.5-pro-preview-05-06}, whereas both our reproductions use the alias \texttt{gemini-2.5-pro} (the current production endpoint). Any per-snapshot behavioral differences propagate uniformly to both rows of Table~\ref{tab:core_metric_comparison}.

\subsection{Details for MedSG-Bench Experiments}


For the implementation phase, we restrict our reproduction to the Visual Patch Grounding (VPG) sub-task as a representative Image Consistency Grounding task and decompose the reproduction into three implementation components:
\begin{itemize}
  \item[\textbf{C1}] \textbf{Data loading.} Loading the \href{https://huggingface.co/datasets/MedSG-Bench/MedSG-Bench}{dataset from HuggingFace} (JSON manifest + zipped image bundles), path-rebasing the image references, and correctly parsing the per-sample fields.
  \item[\textbf{C2}] \textbf{Generation model configuration.} Calling Qwen2.5-VL with the correct hyperparameter (\texttt{max\_new\_tokens}=128, zero-shot, no in-context examples).
  \item[\textbf{C3}] \textbf{Evaluation pipeline.} Parse the model-predicted bounding box in the expected format \texttt{(x\_min, y\_min), (x\_max, y\_max)}, handle model-specific coordinate conventions when needed, and compute the paper's two evaluation metrics: \textsc{IoU} and \textsc{Acc@0.5}.
\end{itemize}

For the MedSG-Bench benchmark, the \texttt{PDF$\rightarrow$CT} stage succeeded in a single run: the task skeleton and  \texttt{TaskSolution} artifacts were correctly populated at iteration~$1$, with $23/24$ verifiable field categories filled ($\approx{}95.8\%$). The only missing category was the per-baseline decoding hyperparameters, which are not specified in the paper. We therefore track implementation completeness as a function of the number of human guiding prompts~$k$: at $k{=}0$, the agent correctly implemented \textbf{C1} and \textbf{C3} in both pipelines, covering the VPG data-loading path and the \textsc{IoU}/\textsc{Acc@0.5} evaluation logic, for an initial accuracy of $2/3$ ($67\%$). \textbf{C2} required one additional guiding prompt because the paper does not specify the per-baseline decoding hyperparameters needed for exact reproduction. After $k{=}1$, once the missing decoding configuration was supplied from the authors' evaluation script \href{https://github.com/Yuejingkun/MedSG-Bench/blob/main/eval/MedSG_Bench.py}{\texttt{eval/MedSG\_Bench.py}}, both pipelines reached $3/3$ ($100\%$). This suggests that the remaining gap reflects discrepancies in the agent-generated data-loading and response-parsing implementation, rather than model choice, prompting strategy, or benchmark-specification drift. Table~\ref{tab:medsgbench_metric_comparison_ci} compares the paper-reported metrics with our implementation. Overall, our implementation reproduces the paper-reported results to a comparable level across most baselines. The main exception is \texttt{Qwen2.5-VL-72B}, where performance degrades primarily because many predictions fail to match the expected bounding-box output format and are therefore not parsed correctly by the evaluation pipeline.

\paragraph{Reproduction setup and compute resources.}
We consider two input conditions, \texttt{CT$\rightarrow$code} and \texttt{PDF$\rightarrow$code}, and evaluate the generated implementations on Task~6 using the Qwen2.5-VL Instruct family. All inference experiments were run on NVIDIA H100~80\,GB GPUs, 12 CPU cores, bf16 precision, batch size $1$, and deterministic decoding using \texttt{do\_sample{=}False} and \texttt{max\_new\_tokens{=}128}. Per-cell wall time was measured over the full $n{=}1000$ VPG split. Table~\ref{tab:medsg_compute} reports the wall time for each pipeline.

\begin{table*}[t]
\centering
\setlength{\tabcolsep}{5pt}
\renewcommand{\arraystretch}{1.15}
\resizebox{\linewidth}{!}{
\begin{tabular}{llccccc}
\toprule
\textbf{Model} & \textbf{Metric}
& \textbf{Reported} 
& \multicolumn{2}{c}{\textbf{PDF$\rightarrow$CT$\rightarrow$code}}
& \multicolumn{2}{c}{\textbf{PDF$\rightarrow$code (direct)}} \\
\cmidrule(lr){4-5} \cmidrule(lr){6-7}
& & \textbf{(\%)} & \textbf{Obtained [95\% CI]} & \textbf{$\Delta$}
& \textbf{Obtained [95\% CI]} & \textbf{$\Delta$} \\
\midrule

\multirow{2}{*}{Qwen2.5-VL-3B}
& IoU     & 27.36 & 26.83 [26.10, 27.50] & -0.53 & 27.22 [26.50, 27.91] & -0.14 \\
& Acc@0.5 &  3.40 &  3.10 [2.10, 4.10]   & -0.30 &  3.20 [2.10, 4.30]   & -0.20 \\
\midrule

\multirow{2}{*}{Qwen2.5-VL-7B}
& IoU     & 28.87 & 30.17 [29.26, 31.12] & +1.30 & 30.81 [29.88, 31.71] & +1.94 \\
& Acc@0.5 &  5.70 &  9.70 [7.80, 11.70]   & +4.00 &  9.80 [7.90, 11.60]  & +4.10 \\
\midrule

\multirow{2}{*}{Qwen2.5-VL-32B}
& IoU     & 26.92 & 27.20 [26.29, 28.14] & +0.28 & 27.20 [26.24, 28.16] & +0.28 \\
& Acc@0.5 &  7.10 &  7.80 [6.30, 9.50]   & +0.70 &  7.80 [6.20, 9.50]   & +0.70 \\
\midrule

\multirow{2}{*}{Qwen2.5-VL-72B}
& IoU     & 26.45 & 14.39 [13.29, 15.43] & -12.06 & 28.08 [27.13, 28.89] & +1.63 \\
& Acc@0.5 &  6.30 &  2.00 [1.20, 2.90]    & -4.30  &  7.00 [5.60, 8.60]   & +0.70 \\
\bottomrule
\end{tabular}
}
\caption{Comparison between reported MedSG-Bench results and our reproduced results. Obtained values are reported as bootstrap means with 95\% confidence intervals. $\Delta$ denotes obtained minus reported.}
\vspace{-1em}
\label{tab:medsgbench_metric_comparison_ci}
\end{table*}

\begin{table}[H]
\centering
\begin{tabular}{@{}lrrr@{}}
\toprule
\textbf{Model} & \textbf{\texttt{PDF$\rightarrow$code}} & \textbf{\texttt{CT$\rightarrow$code}}  \\
\midrule
Qwen2.5-VL-3B  & $5.2$\,min  & $5.3$\,min  \\
Qwen2.5-VL-7B  & $5.3$\,min  & $5.1$\,min  \\
Qwen2.5-VL-32B & $10.2$\,min & $10.1$\,min \\
Qwen2.5-VL-72B & $18.8$\,min & $18.5$\,min  \\
\bottomrule
\end{tabular}
\vspace{0.5em}
\caption{Per-cell wall time for MedSG-Bench inference on the full VPG split.}
\label{tab:medsg_compute}
\end{table}

\subsection{Details for NOVA Benchmark Experiments}

All the experiments involving the NOVA benchmark were run using \href{https://antigravity.google/}{Google's Antigravity} agentic workflow with Gemini Pro 3.1.

For the NOVA benchmark, the \texttt{PDF$\rightarrow$CT} stage achieved 100\% coverage on the first attempt with no additional guidance, taking the agent approximately ten minutes. 

\paragraph{Analysis of CT-Only Pipeline.}
We now detail the implementation of the NOVA benchmark in the case where only Croissant Tasks files were allowed, but not the paper PDF.

The implementation is organized into three components:
\begin{itemize}
  \item[\textbf{C1}] \textbf{Data loading.} Loading the dataset from HuggingFace (Parquet format) and fetching the referenced brain MRI images directly from the HuggingFace repository.
  \item[\textbf{C2}] \textbf{Generation model configuration.} Calling Gemini 2.5 Flash with task-specific prompts for anomaly localization, image captioning, and differential diagnosis, using a fixed \texttt{temperature}=0.1 and \texttt{max\_output\_tokens}=4096.
  \item[\textbf{C3}] \textbf{Evaluation pipeline.} Parsing predictions across three sub-tasks: computing COCO metrics (mAP, ACC50) for Localization; BLEU-4, METEOR, and custom keyword F1 for Captioning; and Top-1/Top-5 scores for Reasoning, followed by calculating 95\% confidence intervals via bootstrapping with 1000 iterations.
\end{itemize}

We track the fraction of metrics correctly produced by the agent, as well as whether the agent noticed their existence or not, as a function of the number of human guiding prompts~$k$. Please note that in the main paper we report the number of metrics correctly implemented over the total number of metrics, while here we take into account whether the agent took the existence of metrics and subTasks into account; this is a slightly different rubric. We note that we deliberately omitted the implementation of the BERT-score metric due to environment-specific development restrictions.
At $k{=}0$ (zero guidance prompts), the agent worked for nine minutes and attempted to generate all metrics. However, it failed to use available specialized libraries like \texttt{pycocotools} and \texttt{evaluate}, leading to incorrect or proxy implementations for the \textsc{mAP} metrics, Binary \textsc{F1}, and Top-1/Top-5 accuracies. This yielded an initial accuracy of approximately 77.4\% of the target implementation.
After $k{=}1$ guiding prompt, instructing the agent to leverage the available libraries and fix the proxy implementations, the agent worked for about ten minutes. This successfully corrected the \textsc{mAP} implementations, bringing the completion rate (existence + correctness) to 87.1\%. However, the Binary \textsc{F1} and diagnostic accuracies still required further refinement.
Finally, after $k{=}2$ guiding prompts, which detailed the exact computation of the Binary \textsc{F1} and Top-1/Top-5 metrics along with output post-processing, the agent worked for twenty minutes.

By this point, out of 14 metrics considered (excluding BERT F1, Coverage, and Entropy), 12 were found to be correctly implemented ($85.7\%$ correctness). Specifically, metrics for anomaly localization such as mAP@30, mAP@50, mAP@[50:95], TP30, FP30, and FNR are correctly implemented using standard tools or formulas. For image description, BLEU-4, METEOR, and binary classification metrics are correctly calculated, although the latter rely on a Large Language Model (LLM) for ground truth extraction. For diagnostic reasoning, Top-1 and Top-5 accuracies are correctly implemented using LLM-based semantic matching as described in the paper. However, we identified discrepancies in three metrics: detection accuracy (ACC50) is implemented as a dataset-level Jaccard index rather than standard accuracy, and the Clinical F1 and Modality F1 scores are calculated via set operations on the union of all keywords across the dataset, failing to account for per-image associations.

\paragraph{Analysis of PDF-Only Pipeline.}
The implementation is organized into three implementation components to handle the three sub-tasks (Localization, Captioning, and Reasoning):
\begin{itemize}
  \item[\textbf{C1}] \textbf{Data loading.} Loading the \href{https://huggingface.co/datasets/c-i-ber/Nova}{dataset from HuggingFace} and correctly parsing both the HuggingFace dataset images and the local ground truth CSV files containing bounding boxes, captions, and case metadata.
  \item[\textbf{C2}] \textbf{Generation model configuration.} Calling Gemini 2.5 Flash via REST API with the correct hyperparameters (\texttt{temperature}=0.1, \texttt{maxOutputTokens}=2048). The implementation handles three distinct tasks: zero-shot localization (requesting normalized coordinates), image description (using a specific radiologist system prompt), and diagnostic reasoning (text-to-text based on clinical history and MRI findings).
  \item[\textbf{C3}] \textbf{Evaluation pipeline.} Processing evaluation for the three sub-tasks: computing COCO metrics and manual \textsc{Acc@0.5} for localization; computing BLEU-4, METEOR, and LLM-judged keyword set-based \textsc{F1} for captioning; and using an LLM-as-a-judge to compute Top-1 and Top-5 accuracy for diagnostic reasoning.
\end{itemize}


We identify significant discrepancies in the implementation of several metrics. Out of the $14$ metrics analyzed (ignoring BERT F1, Coverage, and Entropy, which were not implemented), only $7$ were found to be correctly implemented according to the paper's specifications, resulting in an overall correct implementation fraction of $50\%$. 
Specifically, standard metrics computed via established libraries such as mAP@50 and mAP@[50:95], as well as TP30, FP30, BLEU-4, METEOR, and the macro Binary F1 score were correctly implemented. However, several metrics deviated from the paper's description: mAP@30 was incorrectly calculated as an F1 score rather than actual Average Precision; FNR was computed at an IoU threshold of $0.5$ instead of the $0.3$ threshold implied by the text; the clinical and modality keyword F1 scores were calculated on the pooled unique vocabulary across the entire dataset rather than on an instance-by-instance basis; and the diagnostic reasoning accuracies (Top-1 and Top-5) relied on Gemini for semantic matching instead of the specified GPT-4o. Furthermore, the ACC50 metric was implemented as a Jaccard index (Threat Score) rather than standard classification accuracy.

Table~\ref{tab:nova_metric_comparison} summarizes the discrepancies between the results reported in the original publication and those obtained via our agentic implementations. It is important to note that a correct metric implementation does not necessarily yield numerical scores identical to the published values, as exemplified by the Binary F1 score. This variance can stem from several confounding factors outside the scope of the specification, including the use of different model versions (Gemini 2.0 Flash vs. Gemini 2.5 Flash), potential inaccuracies in the paper's original script, or under-specified evaluation conditions that lead to diverging interpretations of the same metric.

\begin{table}[htbp]
\centering
\begin{tabular}{@{}lccccc@{}}
\toprule
\textbf{Metric} & \textbf{Reported} & \multicolumn{2}{c}{\textbf{PDF Only}} & \multicolumn{2}{c}{\textbf{Croissant Tasks Only}} \\
\cmidrule(lr){3-4} \cmidrule(lr){5-6}
 & \textbf{(\%)} & \textbf{Obtained (\%)} & \textbf{$\Delta$ (\%)} & \textbf{Obtained (\%)} & \textbf{$\Delta$ (\%)} \\
\midrule
\multicolumn{6}{c}{\textbf{Anomaly Localization}} \\ \midrule
mAP@30 & 20.16 & \textcolor{red!70!black}{35.86 {\scriptsize [33.08, 38.71]}} & \textcolor{red!70!black}{+15.70} & \textcolor{green!60!black}{6.07 {\scriptsize [4.94, 7.29]}} & \textcolor{green!60!black}{-14.09} \\
mAP@50 & 7.37 & \textcolor{green!60!black}{3.05 {\scriptsize [2.18, 4.12]}} & \textcolor{green!60!black}{-4.32} & \textcolor{green!60!black}{1.59 {\scriptsize [1.06, 2.37]}} & \textcolor{green!60!black}{-5.78} \\
mAP@[50:95] & 1.99 & \textcolor{green!60!black}{0.79 {\scriptsize [0.53, 1.30]}} & \textcolor{green!60!black}{-1.20} & \textcolor{green!60!black}{0.45 {\scriptsize [0.27, 0.90]}} & \textcolor{green!60!black}{-1.54} \\
ACC50 & 8.83 & \textcolor{red!70!black}{8.52 {\scriptsize [7.28, 9.69]}} & \textcolor{red!70!black}{-0.31} & \textcolor{red!70!black}{5.42 {\scriptsize [4.71, 6.16]}} & \textcolor{red!70!black}{-3.41} \\
FNR & 78.70 & \textcolor{red!70!black}{81.93 {\scriptsize [79.66, 84.21]}} & \textcolor{red!70!black}{+3.23} & \textcolor{green!60!black}{92.09 {\scriptsize [91.09, 93.08]}} & \textcolor{green!60!black}{+13.39} \\ \midrule
\multicolumn{6}{c}{\textbf{Image Description}} \\ \midrule
BLEU-4 & 1.83 & \textcolor{green!60!black}{1.71 {\scriptsize [1.51, 1.91]}} & \textcolor{green!60!black}{-0.12} & \textcolor{green!60!black}{1.64 {\scriptsize [1.44, 1.84]}} & \textcolor{green!60!black}{-0.19} \\
METEOR & 15.20 & \textcolor{green!60!black}{18.19 {\scriptsize [17.57, 18.80]}} & \textcolor{green!60!black}{+2.99} & \textcolor{green!60!black}{18.11 {\scriptsize [17.43, 18.84]}} & \textcolor{green!60!black}{+2.91} \\
Modality Term F1 & 59.80 & \textcolor{red!70!black}{92.10 {\scriptsize [85.71, 100.00]}} & \textcolor{red!70!black}{+32.30} & \textcolor{red!70!black}{95.84 {\scriptsize [87.50, 100.00]}} & \textcolor{red!70!black}{+36.04} \\
Clinical Term F1 & 19.80 & \textcolor{red!70!black}{35.34 {\scriptsize [33.99, 36.69]}} & \textcolor{red!70!black}{+15.54} & \textcolor{red!70!black}{36.30 {\scriptsize [34.60, 38.05]}} & \textcolor{red!70!black}{+16.50} \\
Binary F1 & 5.30 & \textcolor{green!60!black}{89.11 {\scriptsize [87.08, 91.21]}} & \textcolor{green!60!black}{+83.81} & \textcolor{green!60!black}{87.26 {\scriptsize [85.26, 89.30]}} & \textcolor{green!60!black}{+81.96} \\ \midrule
\multicolumn{6}{c}{\textbf{Diagnostic Reasoning}} \\ \midrule
Top-1 Accuracy & 24.20 & \textcolor{red!70!black}{15.50 {\scriptsize [13.34, 17.97]}} & \textcolor{red!70!black}{-8.70} & \textcolor{green!60!black}{19.75 {\scriptsize [17.09, 22.59]}} & \textcolor{green!60!black}{-4.45} \\
Top-5 Accuracy & 38.40 & \textcolor{red!70!black}{21.96 {\scriptsize [19.40, 24.59]}} & \textcolor{red!70!black}{-16.44} & \textcolor{green!60!black}{40.80 {\scriptsize [37.31, 44.12]}} & \textcolor{green!60!black}{+2.40} \\ \bottomrule
\end{tabular}%
\vspace{1em}
\caption{Comparison of NOVA benchmark results (Gemini 2.0 Flash) against obtained results (Gemini 2.5 Flash) under two information settings: utilizing only the paper's PDF versus using only Croissant Tasks files. The 95\% confidence intervals are reported in brackets, computed via bootstrapping with $N{=}1000$ iterations. Values in green indicate metrics that were correctly implemented according to the paper's specifications, while values in red indicate implementations that deviated from the paper.}
\vspace{-1em}
\label{tab:nova_metric_comparison}
\end{table}

\subsection{Details for SAGE-Eval Experiments}

\begin{table}[htbp]
    \centering
    \resizebox{\textwidth}{!}{%
    \begin{tabular}{@{}lccccc@{}}
        \toprule
        \textbf{Metric} & \textbf{Reported} & \multicolumn{2}{c}{\textbf{PDF Only}} & \multicolumn{2}{c}{\textbf{Croissant Tasks Only}} \\
        \cmidrule(lr){3-4} \cmidrule(lr){5-6}
        & \textbf{(\%)} & \textbf{Obtained (\%)} & \textbf{$\Delta$ (\%)} & \textbf{Obtained (\%)} & \textbf{$\Delta$ (\%)} \\
        \midrule
        \multicolumn{6}{c}{\textbf{SAGE-Eval}} \\
        \midrule
        Model-level Safety Score & 30.77 {\scriptsize [26.24, 35.30]} & \textcolor{green!60!black}{32.69 {\scriptsize [26.08, 40.00]}} & \textcolor{green!60!black}{+1.92} & \textcolor{green!60!black}{34.62 {\scriptsize [27.87, 41.54]}} & \textcolor{green!60!black}{+3.85} \\
        Area under Safety Curve (AUC) & 68.56 & \textcolor{green!60!black}{71.69 {\scriptsize [67.79, 75.66]}} & \textcolor{green!60!black}{+3.13} & \textcolor{green!60!black}{71.69 {\scriptsize [67.51, 75.93]}} & \textcolor{green!60!black}{+3.13} \\
        \bottomrule
    \end{tabular}%
    }
    \vspace{1em}
    \caption{Comparison of SAGE-Eval benchmark results (Gemini 2.0 Flash) against obtained results under two information settings: direct implementation from the paper PDF vs.\ implementation from the Croissant Task description. 95\% confidence intervals for the obtained metrics are reported in brackets, computed via fact-level bootstrapping with $N{=}1000$ iterations to account for the grouped structure of the safety facts. Values in green indicate that the metrics were correctly implemented according to the paper's specifications (e.g., matching the threshold-based AUC formula and judge prompt structure).}
    \vspace{-1em}
    \label{tab:sageeval_metric_comparison}
\end{table}

Involving the SAGE-Eval benchmark, all the experiments were run using \href{https://antigravity.google/}{Google's Antigravity} agentic workflow with Gemini Pro 3.1. 

The \texttt{PDF$\rightarrow$CT} stage achieved 100\% accuracy with no additional guidance: all 13 verifiable field categories across the \texttt{TaskProblem} and 15 \texttt{TaskSolution} skeletons were correctly populated on the first attempt. 
\old{During this stage, the agent also self-identified three upstream SHACL shape bugs in \texttt{croissant-tasks-shapes.ttl} and attempted to patch them by using a bundled structural workaround (\texttt{infra/\_structural\_check.py}).}

Separately, for the implementation phase, we decompose the benchmark reproduction into three implementation components:
\begin{itemize}
  \item[\textbf{C1}] \textbf{Data loading.} Loading the \href{https://huggingface.co/datasets/YuehHanChen/SAGE-Eval}{dataset from HuggingFace} and correctly parsing the relevant columns (\texttt{prompt}, \texttt{safety\_fact}, \texttt{category}).
  \item[\textbf{C2}] \textbf{Generation model configuration.} Calling the target LLM with the correct hyperparameters (\texttt{temperature}=0, \texttt{max\_tokens}=2000).
  \item[\textbf{C3}] \textbf{Evaluation pipeline.} Applying the correct judge prompt template and hyperparameters (\texttt{max\_tokens}=300).
\end{itemize}
We then track the fraction correctly produced by the agent as a function of the number of human guiding prompts~$k$. At $k{=}0$ (zero guiding prompts), the agent correctly implemented C1 and C2 in both pipelines, yielding an initial accuracy of 2/3 (67\%). However, it produced a \emph{zero-shot} judge prompt with \texttt{max\_tokens}=100. This occurred because the paper's appendix describes the evaluation as ``LLM-as-a-judge'' without providing the actual few shot examples in the prompt template; the full few-shot template and hyperparameters were only available in the GitHub repository file \href{https://github.com/YuehHanChen/SAGE-Eval/blob/main/src/utils/prompts/eval.py}{\texttt{src/utils/prompts/eval.py}}. In both cases, the final accuracy reached 3/3 (100\%) after $k{=}1$ guiding prompt. We compare the metrics reported in the paper with the metrics obtained in Table~\ref{tab:sageeval_metric_comparison}. All model inferences were performed via the Google Gemini API. Approximately 1.2 seconds for generation and 0.8 seconds for evaluation, resulting in a total processing time of $\sim$2.0 seconds per instance.  Each full evaluation run took approximately 105 minutes.



\end{document}